%% file: 00_main.tex
\documentclass{article} 
\usepackage{aloe2022,times}
\usepackage{graphicx}

\input{math_commands.tex}

\usepackage{hyperref}
\usepackage{url}
\usepackage{bbm}

\title{Streaming Inference for Infinite\\ Non-Stationary Clustering}

\author{Rylan Schaeffer\\
Computer Science\\
Stanford University\\
\texttt{rschaef@cs.stanford.edu} \\
\And
Gabrielle Kaili-May Liu\\
Brain and Cognitive Sciences \\
Massachusetts Institute of Technology\\
\texttt{gkml@mit.edu} \\
\And
Yilun Du\\
Electrical Engineering \& Computer Science\\
Massachusetts Institute of Technology\\
\texttt{yilundu@mit.edu} \\
\AND
Scott Linderman \\
Statistics \& Electrical Engineering\\
Stanford University\\
\texttt{scott.linderman@stanford.edu}
\And
Ila Rani Fiete \\
Brain and Cognitive Sciences \\
Massachusetts Institute of Technology\\
\texttt{fiete@mit.edu}
}

\iclrfinalcopy 
\begin{document}

\maketitle

\begin{abstract}
Learning from a continuous stream of non-stationary data in an unsupervised manner is arguably one of the most common and most challenging settings facing intelligent agents. Here, we attack learning under all three conditions (unsupervised, streaming, non-stationary) in the context of clustering, also known as mixture modeling. We introduce a novel clustering algorithm that endows mixture models with the ability to create new clusters online, as demanded by the data, in a probabilistic, time-varying, and principled manner. To achieve this, we first define a novel stochastic process called the Dynamical Chinese Restaurant Process (Dynamical CRP), which is a non-exchangeable distribution over partitions of a set; next, we show that the Dynamical CRP provides a non-stationary prior over cluster assignments and yields an efficient streaming variational inference algorithm. We conclude with experiments showing that the Dynamical CRP can be applied on diverse synthetic and real data with Gaussian and non-Gaussian likelihoods.
\end{abstract}




\input{01_introduction}

\input{02_background}

\input{03_method}

\input{04_results}

\input{05_discussion}

\input{06_conclusion}

\clearpage
\bibliography{references}
\bibliographystyle{iclr2022_conference}

\clearpage

\appendix

\input{07_appendix}

\end{document}

%% file: math_commands.tex
\usepackage{amsmath,amsfonts,bm}

\def\eqref#1{equation~\ref{#1}}

\def\1{\bm{1}}

\DeclareMathAlphabet{\mathsfit}{\encodingdefault}{\sfdefault}{m}{sl}
\SetMathAlphabet{\mathsfit}{bold}{\encodingdefault}{\sfdefault}{bx}{n}

\DeclareMathOperator{\Tr}{Tr}
\DeclareMathOperator{\defeq}{\stackrel{\text{def}}{=}}
\DeclareMathOperator{\defpropto}{\stackrel{\text{def}}{\propto}}

%% file: 01_introduction.tex
\section{Introduction}

Biological intelligence operates in a radically different data regime than most artificial intelligence. In particular, biological intelligence must contend with data that is (i) unsupervised, (ii) streaming, and (iii) non-stationary, either as a consequence of the agent, its environment, or both. One goal of lifelong learning is to make artificial intelligence significantly more capable in this data regime, and accomplishing that goal requires asking and answering how agents in this data regime ought to approach learning.

In this paper, we consider the specific unsupervised problem of clustering, also known as mixture modeling. Clustering is a ubiquitous and important problem in its own right, with widespread applications, but clustering can also serve as an intermediary sub-goal in service of other goals: for instance, an agent in a partially observable world may wish to cluster sensory observations into world states to then use for spatial navigation or reinforcement learning. In this paper, we specifically consider an agent who receives a single stream of observations from non-stationary clusters, with no ability to revisit past observations, but must nonetheless identify the clusters and assign observations to them. In this data regime, the number of clusters (i.e. the number of mixture components) is unknown and theoretically could be unbounded, and so the agent must use a clustering algorithm capable of growing in representational capacity as more observations are encountered.

In this paper, we define a novel distribution over partitions of a set that we call the Dynamical Chinese Restaurant Process (Dynamical CRP), due to its relationship with the Chinese Restaurant Process (CRP) \citep{ferguson_bayesian_1973, blackwell_ferguson_1973, antoniak_mixtures_1974}. We then show how the Dynamical CRP can be used as a prior over cluster assignments in a manner that yields an efficient streaming clustering algorithm designed for non-stationary data. Starting with synthetic Gaussian and non-Gaussian data, and moving to more sophisticated real data based on  simultaneous localization and mapping (SLAM), we show that streaming inference using the Dynamical CRP achieves comparable or better performance than many common baselines, especially when the data is non-stationary.


%% file: 02_background.tex
\section{Background}

\subsection{Notation}

We consider a single time series of $D$-dimensional observable variables $o_{1:N}$ ($o_n \in \mathbb{R}^D$) occurring at known times $t_{1:N}$ and corresponding to some latent cluster assignment variables $c_{1:N}$ (i.e. $c_n \in \{1, 2, ... \}$), where $\cdot_{1:N}$ denotes the sequence $(\cdot_1, \cdot_2, ..., \cdot_N)$. Our goal is to infer the latent cluster assignments $c_{1:N}$. Each cluster may have corresponding variables $\{\phi_c\}_{c=1}^C$ (e.g., per-cluster means and covariances) that we might also wish to infer. In the non-stationary setting, the clusters may change over time in a manner that we shall specify.

\subsection{Infinite Clustering via the Chinese Restaurant Process}

In the non-stationary streaming data regime, the number of clusters is unknown and unbounded. Consequently, a useful clustering algorithm should be capable of (a) adding clusters as necessitated by the data, (b) generating predictions of future likely clusters, and (c) changing learnt representations of clusters over time. To meet the first two desiderata, many clustering algorithms use the Chinese Restaurant Process (CRP) or its related Dirichlet Process \citep{ferguson_bayesian_1973, antoniak_mixtures_1974, neal_markov_2000, blei_variational_2006, kulis_revisiting_2012}. The CRP is a single-parameter ($\alpha > 0$) stochastic process that defines a discrete distribution over partitions of a set, making it an applicable prior for cluster assignments. The name CRP arises from a story of a sequence of customers (observations) arriving at a restaurant with an infinite number of tables (clusters), each table with infinite capacity. The first customer $c_1$ sits at the first table, and each subsequent customer $c_n$ sits either at an unoccupied table with probability proportional to $\alpha$ or joins an occupied table with probability proportional to the number of preceding customers at that table. Denoting the number of non-empty tables after the first $n$ customers $C_n \defeq \max (c_1, ..., c_n)$, CRP($\alpha$) defines a conditional distribution for the $n$th customer $c_n$ given the preceding customers $c_{<n}$:
\begin{equation}
\begin{aligned}
    &p^{CRP}(c_{n} = c| c_{<n}, \alpha)
    &\propto \begin{cases}
    \sum_{n' < n} \mathbbm{I}(c_{n'} = c) & \text{if }1 \leq c  \leq C_{n-1} \defeq \max(c_1, ..., c_{n-1})\\
    \alpha & \text{if } c = C_{n-1} + 1\\
    0 & \text{otherwise}
    \end{cases}
    \label{eq:crp_defn}
\end{aligned}
\end{equation}

An example application of the CRP is task-free continual learning \citep{lee_neural_2020}. However, the CRP is ill-suited to streaming data because the CRP's conditional form requires knowing the entire history of cluster assignments; \citet{schaeffer_efcient_2021} showed the CRP can be adapted for streaming data by rewriting the CRP in a recursive form:
\begin{equation}
p^{CRP}(c_n = c|\alpha) \propto \sum_{n' < n} p(c_{n'} = c | \alpha) + \alpha \, p(C_{n-1} = c - 1)
\label{eq:crp_recursion}
\end{equation}

The two-part intuition is that (i) if many observations come from cluster $c$, then the next observation is also likely to come from cluster $c$, and (ii) the probability of more clusters should grow with the number of observations, giving the CRP the capacity to create an ``infinite" number of clusters.






\subsection{Non-stationary Variants of the Chinese Restaurant Process}

Although the CRP is widely used, the CRP has two properties which are inappropriate for non-stationary data. First, the CRP is exchangeable, meaning permuting the order of the data does not affect the probability of the resulting partition. Second, the CRP is consistent, meaning marginalizing out any observation is the same as if the observation never existed. To handle non-stationary data, \cite{zhu_time-sensitive_2005} defined the time-sensitive CRP (tsCRP) by introducing exponential decay:
\begin{equation}
\begin{aligned}
    p^{tsCRP}(c_n = c | c_{<n}, \alpha)
    &\propto \begin{cases}
    \sum_{n'<n} \exp((t_n - t_{n'})/\tau) \mathbbm{I}(c_{n'} = c) & \text{if } 1 \leq c \leq C_{n-1}\\
    \alpha & \text{if } c = C_{n-1} + 1\\
    0 & \text{otherwise}
    \end{cases}\label{eq:tscrp_dfn}
\end{aligned}
\end{equation}


\cite{blei_distance_2011} later defined the distance-dependent Chinese Restaurant Process (ddCRP), which assigns customers to other customers in a (possibly cyclic) directed graph. While  flexible, the ddCRP is impractical for streaming inference because observations can be assigned to future observations, and time/space complexities must be quadratic in the number of observations because the pairwise relationships have no structure and thus must all be remembered.

%% file: 03_method.tex
\section{Methods}

\subsection{Desiderata}

Our goal is to define an efficient streaming inference algorithm for infinite non-stationary clustering. To do this, we define a novel stochastic process over partitions of a set called the \textbf{Dynamical CRP} to use as a prior over cluster assignments. The Dynamical CRP is designed with the following goals:
\begin{itemize}
    \item Like the CRP, the Dynamical CRP can create ``infinite" clusters (albeit upper bounded by the number of observations) and can generate predictions of the next likely clusters.
    \item Unlike the CRP, the Dynamical CRP does not assume the observations are i.i.d., exchangeable or consistent, meaning the Dynamical CRP can model non-stationary data.
    \item Unlike the tsCRP, the Dynamical CRP does not restrict the influence of observation times to exponential decay and can therefore capture a richer class of temporal relationships.
    \item Unlike the ddCRP, the Dynamical CRP admits an efficient streaming inference algorithm, which is critical for practical use by agents with finite memory.
\end{itemize}

The Dynamical CRP thus sits in a ``Goldilocks" zone: more powerful than the CRP or tsCRP, but less powerful than the ddCRP so as to still permit efficient streaming inference.

\subsection{High Level Idea}

The heart of the CRP is the ``table occupancies" $N_{c}(t) \defeq \sum_{n' \leq n}\mathbbm{I}(c_n = c) \mathbbm{I}(t_{n'} \leq  t)$, which are the sufficient statistics of the stochastic process. The Dynamical CRP embeds those table occupancies in a dynamical system to evolve endogenously. By choosing or learning dynamics appropriate for a particular task, the Dynamical CRP gains rich time-dependent priors for cluster assignments.

\subsection{Definition}


Let $\mathcal{H}$ be a Hilbert space and $\tilde{N}(t) \in \mathcal{H}$ contain both the ``pseudo" table occupancies $N_c(t)$ and any desired higher-order temporal derivatives. Fix a linear dynamical system $\ell: \tilde{N} \rightarrow \tilde{N}$ and increment the $c_n$-th table $N_{c_n}$ at time $t_n$ by $1$. As before, define $C_n \defeq \max (c_1, ..., c_n)$. The Dynamical CRP, denoted $D\text{-}CRP(\ell, \alpha)$, is defined as the conditional distribution:
\begin{equation}
    p^{D{\text-}CRP}(c_n = c | c_{< n}, t_{\leq n}, \ell, \alpha) \propto \begin{cases}
    N_c(t_n) & \text{if } 1 \leq c \leq C_{n-1}\\
    \alpha & \text{if } c = C_{n-1} + 1\\
    0 & \text{otherwise}
    \end{cases}\label{eq:DCRP_definition}
\end{equation}

Like the CRP, each customer increments a table's occupancy count, but the tables' occupancies can now change endogenously. We next show the flexibility that the Dynamical CRP provides.

\begin{figure}[h]
\centering
\includegraphics[width=0.48\textwidth]{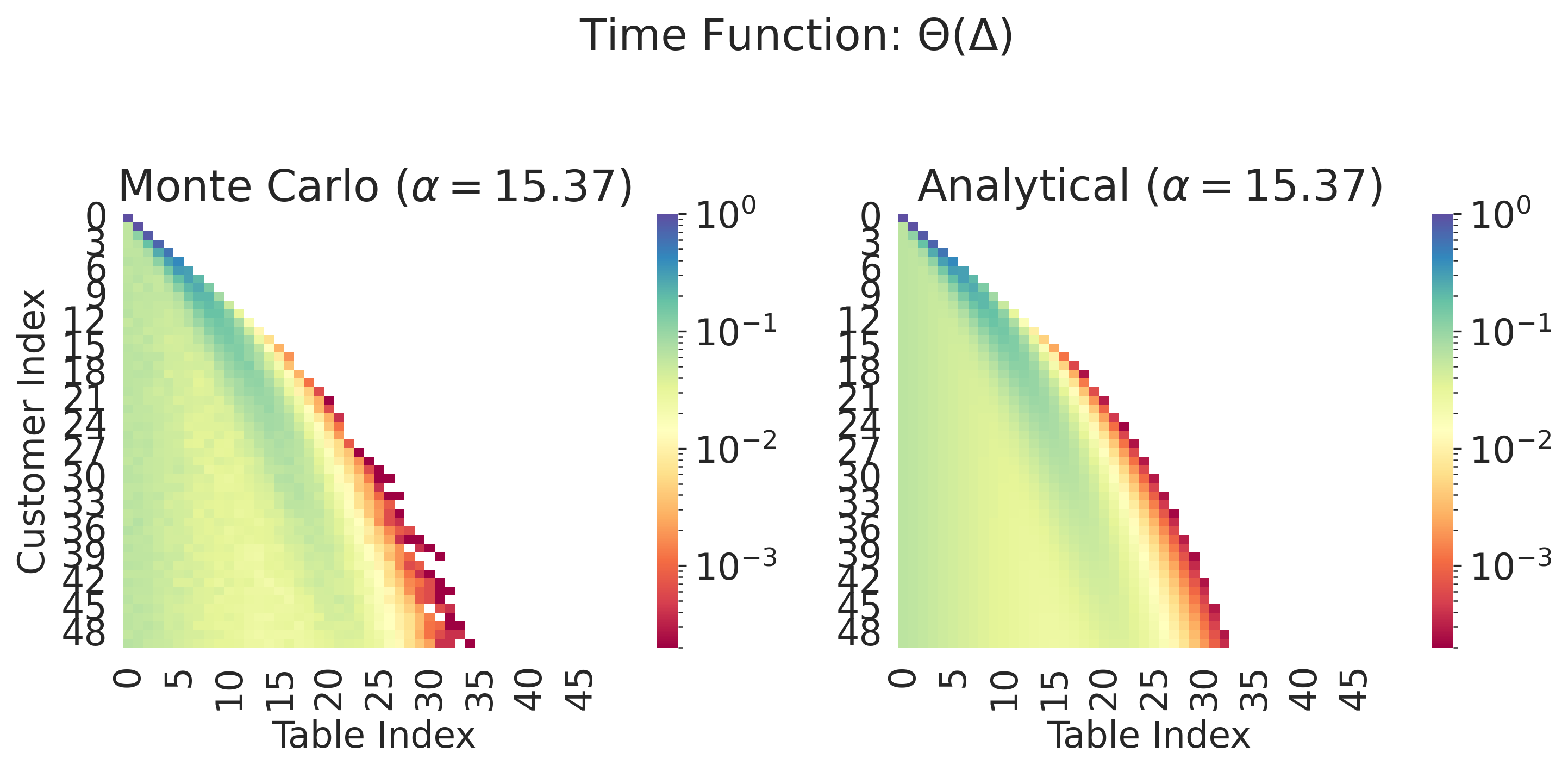}%
\includegraphics[width=0.48\textwidth]{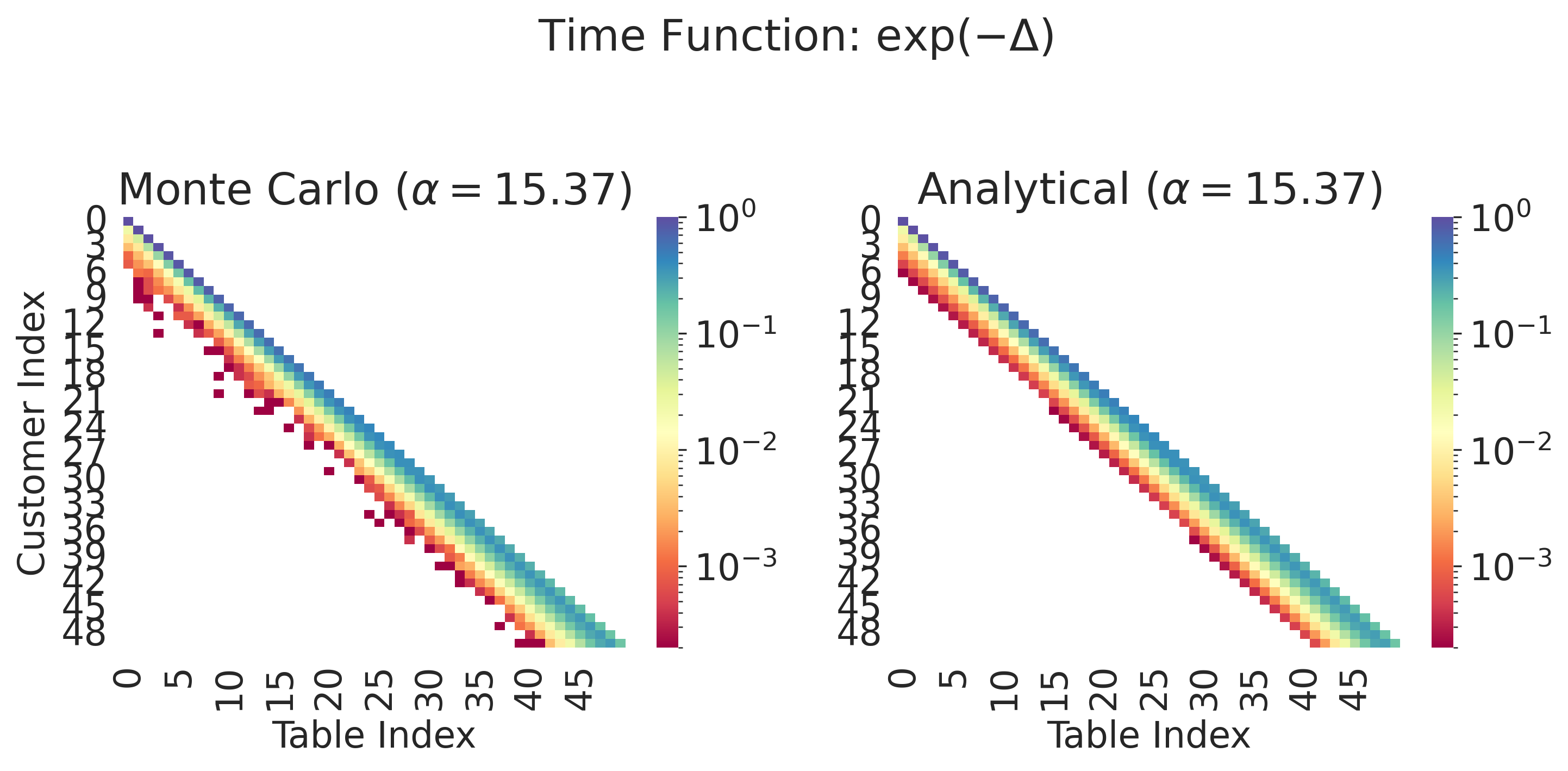}
\includegraphics[width=0.48\textwidth]{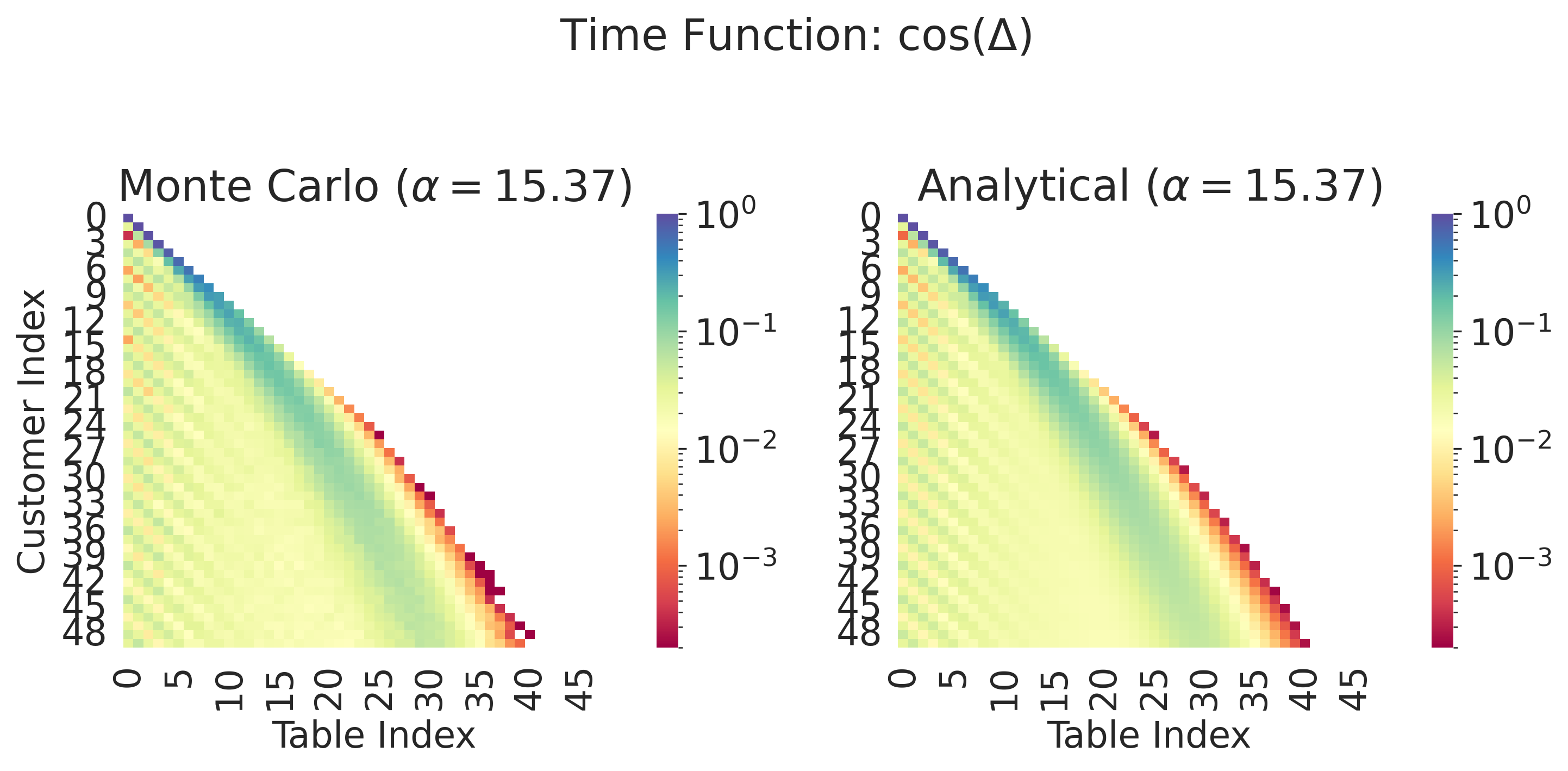}%
\includegraphics[width=0.48\textwidth]{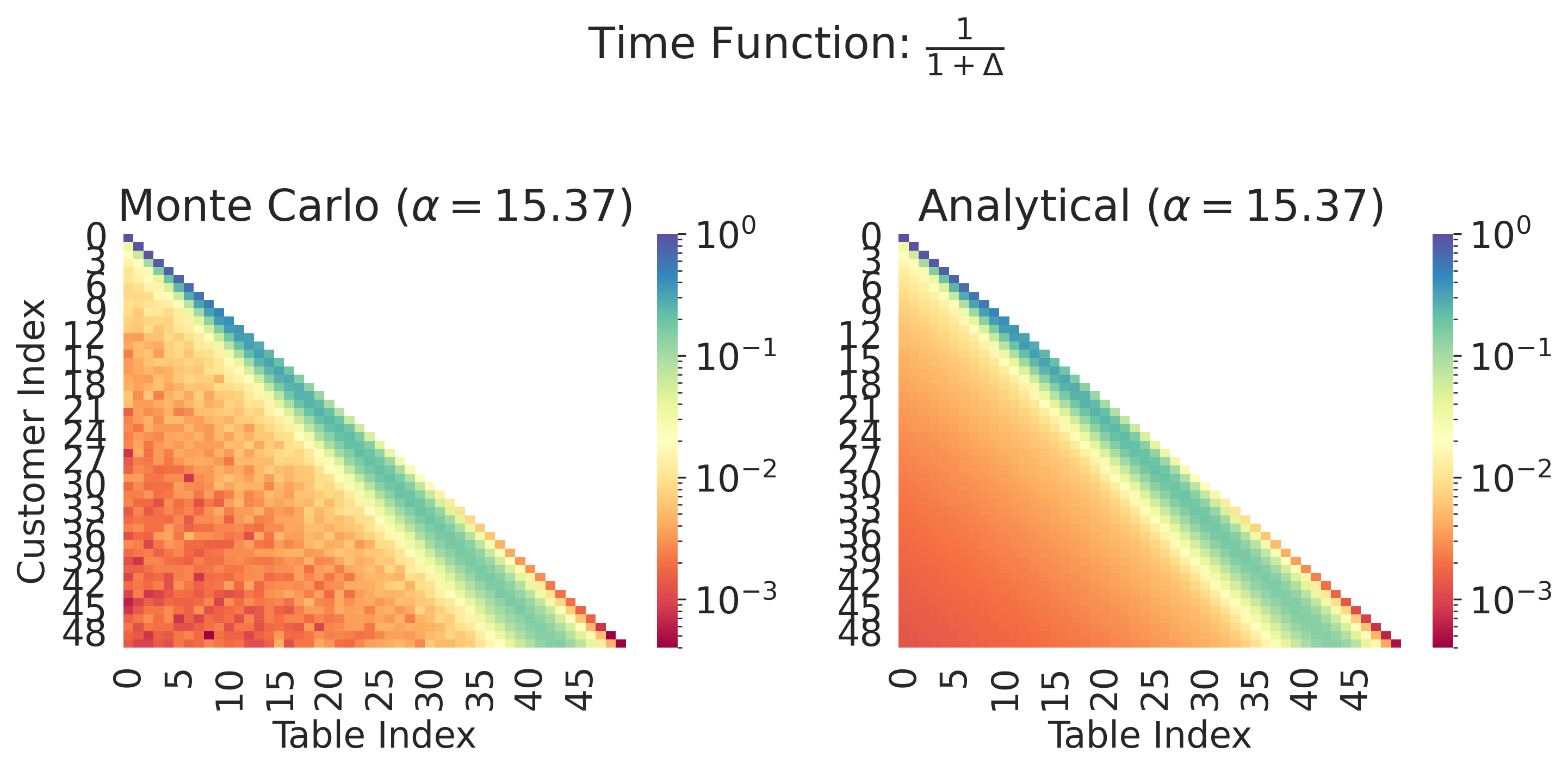}
\caption{The Dynamical Chinese Restaurant Process under 4 different dynamics. Elapsed time is denoted by $\Delta \defeq t_n - t_{n-1}$. Under different dynamics, the Dynamical CRP produces the CRP (Time Function: $\Theta(\Delta)$), the time-sensitive CRP (Time Function: $\exp(-\Delta)$), and new distributions over partitions of a set including sinusoidal (Time Function: $\cos(\Delta))$ and hyperbolic (Time Function: $1/(1+\Delta)$). Columns 1 and 3 are Monte Carlo samples; Columns 2 and 4 are our analytical recursion.}
\label{fig:dcrp_stochastic_process}
\end{figure}

\subsection{Examples}

\textbf{Stationary Dynamics}: Define $\ell(\tilde{N}) \defeq \partial_t \tilde{N}(t)$ with initial conditions $N_c(0) = 0$. Then the Dynamical CRP assumes the data is stationary and simplifies to the CRP (Fig. \ref{fig:dcrp_stochastic_process}, Time Function: $\Theta(\Delta)$).



\textbf{Exponential Dynamics}: Define $\ell(\tilde{N}) \defeq \tau \partial_t \tilde{N}(t) + \tilde{N}(t)$. Then the relevance of previous customers (observations) decays exponentially with elapsed time and the Dynamical CRP simplifies to the time-sensitive CRP (Fig. \ref{fig:dcrp_stochastic_process}, Time Function: $\exp(-\Delta)$).

\textbf{Oscillatory Dynamics}: Suppose we want cluster assignments to be periodic on a particular timescale. For instance, dawn and dusk have visually similar light, but crepuscular animals need to distinguish them. Similarly, fall and spring have similar temperatures, but migratory and  hibernating/aestivating animals need to distinguish them. By defining the dynamics as a linear second order differential equation $\ell(\tilde{N}) \defeq \partial_t^2 \tilde{N}(t) + \tilde{N}(t)$, the Dynamical CRP creates oscillatory table assignments (Fig. \ref{fig:dcrp_stochastic_process}, Time Function: $\cos(\Delta)$).

\textbf{Hyperbolic Dynamics}: Hyperbolic discounting is commonly used in reinforcement learning and observed across species including humans, monkeys, and rats \citep{sozou_hyperbolic_1998, fedus_hyperbolic_2019}. The Dynamical CRP also enables hyperbolic clustering (Fig. \ref{fig:dcrp_stochastic_process}, Time Function: $1/(1 + \Delta)$).




\subsection{Generative Model}

We now define the generative model for the streaming data $o_{1:N}$, observed at known times $t_{1:N}$, using the Dynamical CRP as a prior over cluster assignments $c_{1:N}$:
\begin{equation}
\begin{aligned}
    c_{1:N} \lvert t_{1:N} &\sim D{\text-}CRP(\ell, \alpha)\\
    \phi_k &\sim_{i.i.d.} p(\phi)\\
    o_n \lvert c_n, \{\phi_k\}_{k=1}^{\infty} &\sim p(o_n; \phi_{c_n}) 
\end{aligned}
\label{eq:dcrp_generative_model}
\end{equation}

\subsection{Streaming Inference}

Our approach will be to first show that the Dynamical CRP can be expressed in a recursive form designed for streaming inference, similarly to the CRP, and then use this recursive form to define a variational family for streaming inference.

\subsubsection{Recursive Form of the Dynamical CRP}

As with the CRP, the Dynamical CRP's conditional distribution renders each cluster assignment $c_n$ dependent on the entire history of previous cluster assignments $c_{<n}$. This handicaps its applicability to streaming data. We overcome this by converting the conditional distribution to a marginal distribution by taking the average over all possible histories of cluster assignments (termed \textit{sample paths} in the stochastic processes literature). Omitting $t_{1:N}$ for brevity, 
\begin{align*}
    p(c_{n} = c|\ell, \alpha) &= \mathbb{E}_{p(c_n|\ell, \alpha)}[\mathbbm{I}(c_n = c)]\\
    &= \mathbb{E}_{p(c_{< n} | \ell, \alpha)}[\mathbb{E}_{p(c_n | c_{<n}, \ell, \alpha)}[\mathbbm{I}(c_n = c)]]\\
    &= \mathbb{E}_{p(c_{< n}|\ell, \alpha)}[p(c_n = c | c_{<n}, \ell, \alpha)]
\end{align*}

Substituting the Dynamical CRP's conditional distribution and taking a first-order Taylor series approximation of the expectation yields:
\begin{align*}
    p(c_{n} = c|\ell, \alpha) &= \mathbb{E}_{p(c_{< n}|\ell, \alpha)} \Bigg[ \frac{N_c(t_n)}{\alpha + \sum_c N_c(t_n)} + \frac{\alpha}{\alpha + \sum_c N_c(t_n)} \mathbbm{I}(c = C_{n-1} + 1) \Bigg]\\
    &\approx \frac{\mathbb{E}[N_{c}(t_n)]}{\alpha + \mathbb{E}[\sum_c  N_c(t_n)]} +  \frac{\alpha}{\alpha+\mathbb{E}[\sum_c N_c(t_n)]} p(C_{n-1} = c - 1)
\end{align*}

Abusing notation slightly, we can write $N_c(t_n) = \sum_{n' < n} \ell(\mathbbm{I}(c_{n'} = c), t_{n'}, t_n)$, where $\ell(\cdot, t_{n'}, t_n)$ means advancing the dynamical system from time $t_{n'}$ to time $t_n$. Because both the dynamics $\ell$ and the expectation are linear operators, the two commute and the expectation can be pulled inside: 
\begin{equation*}
    \mathbb{E}[N_c(t)] = \sum_{n': t_{n'} < t} \ell(\mathbb{E}[\mathbbm{I}(c_{n'} = c)], t_{n'}, t) = \sum_{n': t_{n'} < t} \ell(p(c_{n'} = c|\ell, \alpha), t_{n'}, t)
\end{equation*}

\begin{figure}[b]
\centering
\includegraphics[width=0.24\textwidth]{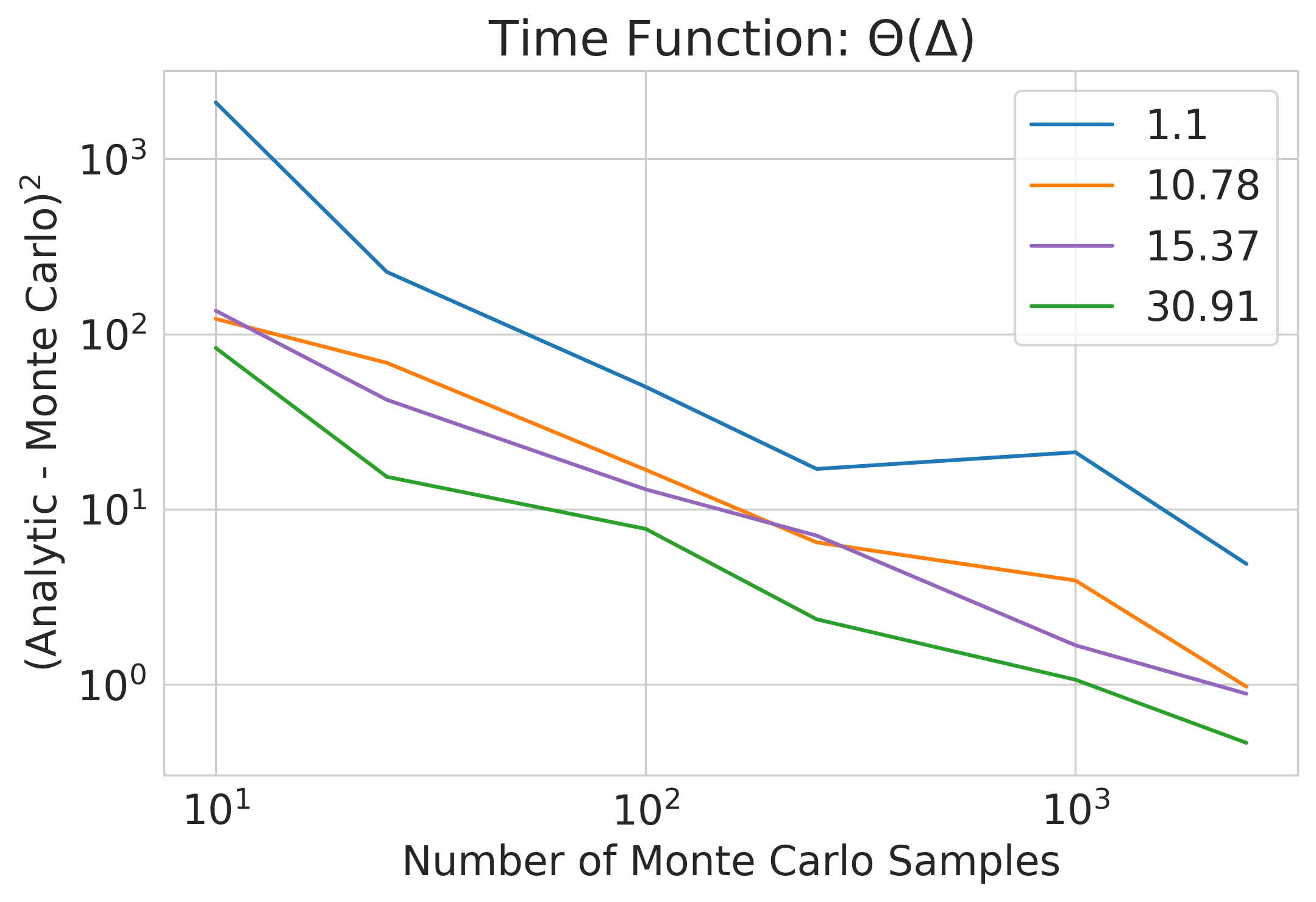}%
\includegraphics[width=0.24\textwidth]{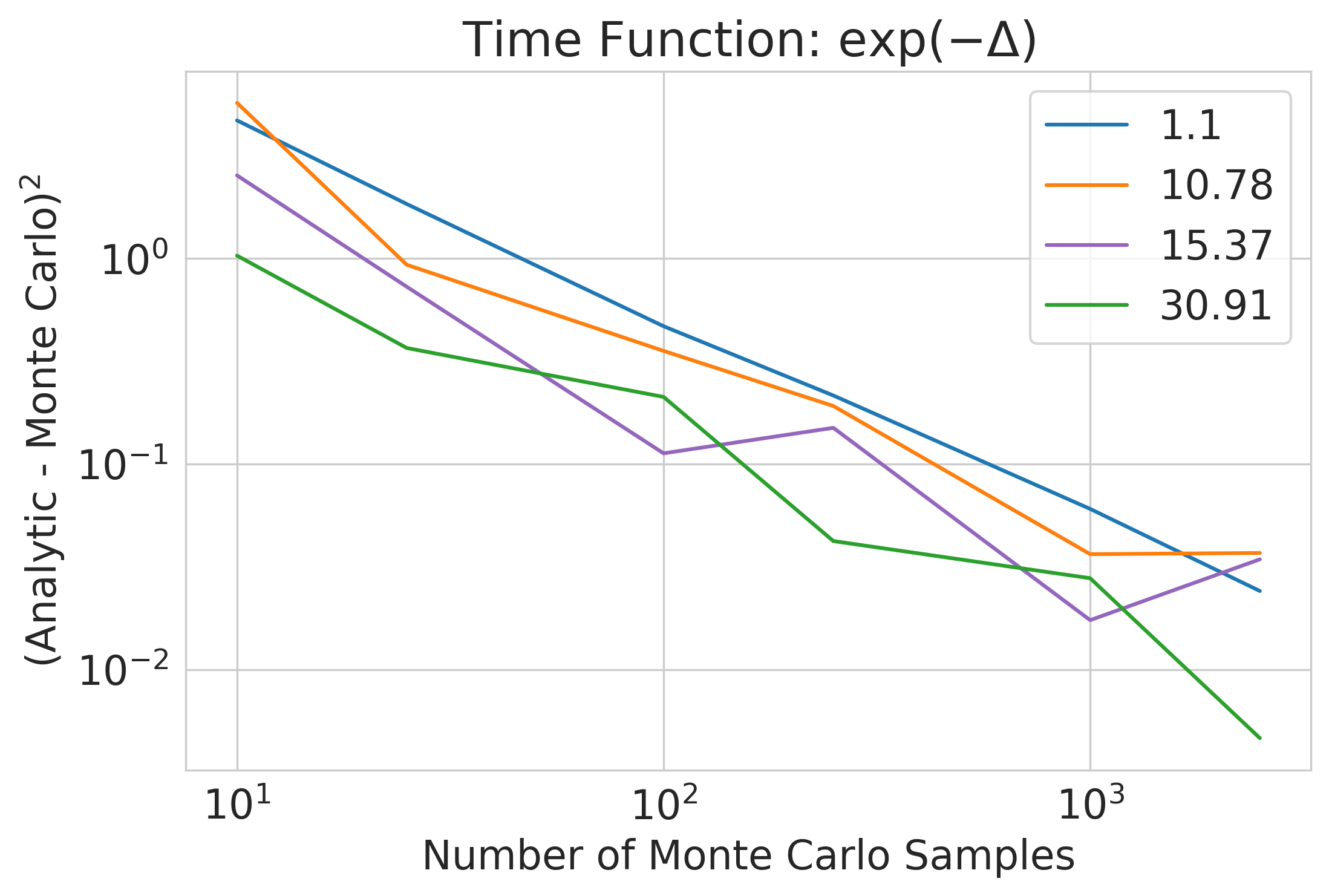}%
\includegraphics[width=0.24\textwidth]{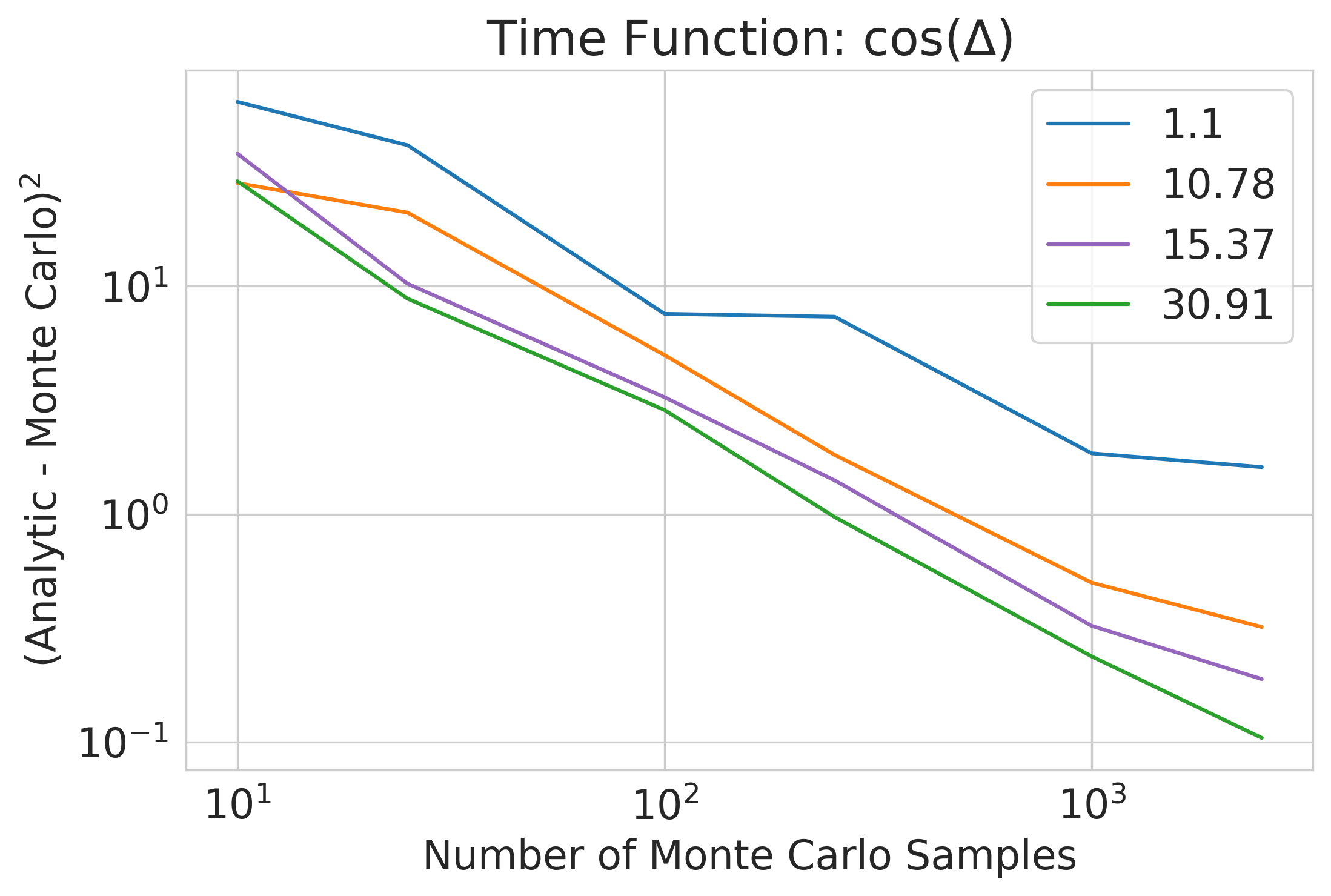}%
\includegraphics[width=0.24\textwidth]{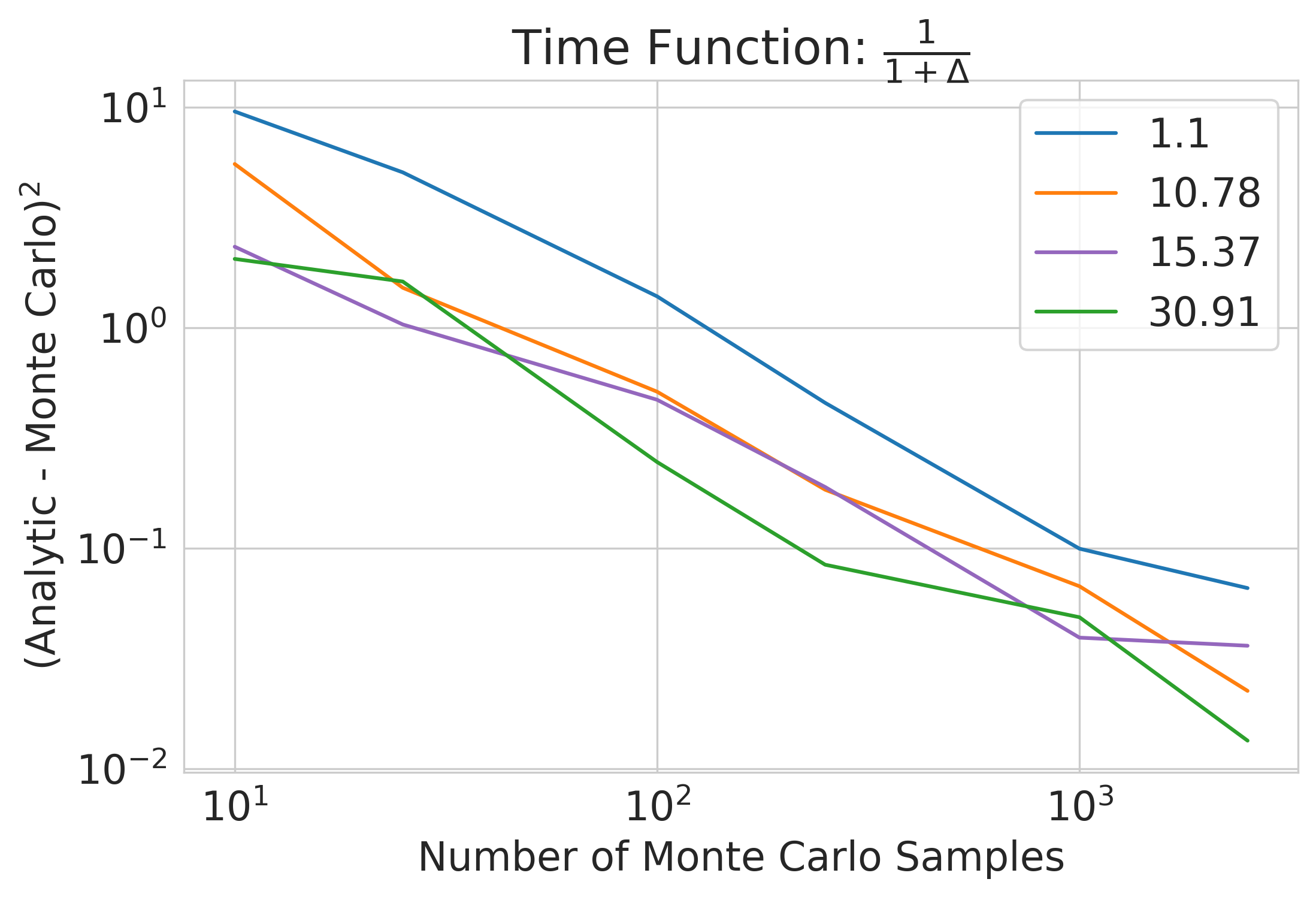}
\caption{\textbf{Mean-Squared Error between analytical expression for $p(c_n|\ell, \alpha)$ and a Monte Carlo estimate.} Over a wide range of $\alpha$ values, the mean-squared error between our analytical expression and Monte Carlo estimates falls approximately as a power law, showing the exactness of Eqn. \ref{eq:dynamical_crp_recursion}.}
\label{fig:dcrp_monte_carlo_vs_analytical}
\end{figure}

Together, this yields the recursive form of the Dynamical CRP:

\begin{equation}
    p(c_{n} = c|\ell, \alpha) \propto \sum_{n': t_{n'} < t} \ell(p(c_{n'} = c|\ell, \alpha), t_{n'}, t_n) +  \alpha \, p(C_{n-1} = c - 1)
    \label{eq:dynamical_crp_recursion}
\end{equation}

The Dynamical CRP's recursive form (Eqn. \ref{eq:dynamical_crp_recursion}) has similar intuition to the CRP's recursive form (Eqn. \ref{eq:crp_recursion}): previous cluster assignments influence the current cluster assignment and clusters can appear with new observations. The key modification is that previous probability masses can now change over time. We confirm the correctness of Eqn. \ref{eq:dynamical_crp_recursion} by comparing the analytical expression to 5000 Monte Carlo samples drawn from the Dynamical CRP's conditional distribution over $\alpha \in \{1.1, 10.78, 15.37, 30.91\}$ and with step, exponential, sinusoidal, and hyperbolic dynamics (Fig. \ref{fig:dcrp_stochastic_process}); visually, the analytical and Monte Carlo plots display excellent agreement. Quantitatively, the mean squared error between the analytical expression for all $p(c_{n}|\ell, \alpha)$ and the Monte Carlo estimates falls approximately as a power law in the number of Monte Carlo samples (Fig. \ref{fig:dcrp_monte_carlo_vs_analytical}) for all $\alpha$ values. This supports our claim that the recursive form of the Dynamical CRP is highly accurate.

\subsubsection{Streaming Inference via Recursive Form of Dynamical CRP}

To perform streaming inference, we start by considering the streaming evidence lower bound: 
\begin{align*}
    p(o_n | o_{<n}) &\geq \mathbb{E}_{q(c_n, \{\phi\}|o_{\leq n};\theta_n)}[\log p(o_n, c_n, \{\phi\} | o_{<n})] + H[q(c_n, \{\phi\},o_{\leq n})]\\
    &= \mathbb{E}_{q(c_n, \{\phi\}|o_{\leq n};\theta_n)}[\log p(o_n| c_n, \{\phi\}, o_{<n}) + \log p(c_n, \{\phi\}|o_{<n})] + H[q(c_n, \{\phi\}|o_{\leq n})]
    \label{eq:true_lower_bound}
\end{align*}

with variational parameters $\theta_n$. However, computing this evidence lower bound is tricky because the filtering prior $p(c_n, \{\phi\} |o_{<n})$ is unknown. Using the recursive form of the Dynamical CRP as inspiration, we replace the filtering prior with an approximate filtering prior:
\begin{align*}
    q(c_n|o_{<n}) &\defpropto \sum_{n' < n} \ell(q(c_{n'} = c|o_{\leq n'}, \ell, \alpha), t_{n'}, t_n) +  \alpha \, q(C_{n-1} = c - 1 | o_{\leq n-1})\\
    q(\{\phi\} |o_{< n}) &\defeq \prod_k q(\phi_k |o_{\leq n-1})\\
    q(c_n, \{\phi\} |o_{<n}) &\defeq q(c_n|o_{<n}) q(\{\phi\} |o_{< n})
\end{align*}

Substituting the approximate filtering prior yields an approximate filtering evidence lower bound that we maximize:
\begin{equation}
    \mathbb{E}_{q(c_n, \{\phi\}|o_{\leq n};\theta_n)}[\log p(o_n| c_n, \{\phi\}, o_{<n}) + \log q(c_n, \{\phi\}|o_{<n})] + H[q(c_n, \{\phi\}|o_{\leq n})]
    \label{eq:approx_lower_bound}
\end{equation}


%% file: 04_results.tex
\section{Experimental Results}

\subsection{Synthetic Mixture of Gaussians}

Following previous work \citep{kulis_revisiting_2012}, we started with synthetic mixtures of Gaussians. We generated datasets of 1000 observations by placing a Gaussian prior on the cluster means $p(\phi) = \mathcal{N}(0, \rho^2 I)$ and sweeping over dynamics, alpha, signal-to-noise ratios, and number of dimensions to construct 3600 datasets total. We compared Dynamical CRP against seven baseline inference algorithms; three are streaming and four are not. The non-streaming algorithms have unfettered access to all observations and therefore serve as upper bounds on performance; any comparison against these non-streaming baselines maximally disfavors our method. The baselines are:
\begin{itemize}
    \item Collapsed Gibbs Sampling (non-streaming) \citep{neal_markov_2000}.
    \item Variational Bayes Dirichlet-Process Gaussian Mixture Model (non-streaming) \citep{blei_variational_2006}, implemented in scikit-learn \citep{pedregosa_scikit-learn_2011}.
    \item K-Means (both streaming and non-streaming variants) \citep{macqueen_methods_1967, lloyd_least_1982}. For both variants, K-Means is given the ground-truth number of clusters.
    \item DP-Means (both streaming and non-streaming variants) \citep{kulis_revisiting_2012, broderick_mad-bayes_2013}.
    \item Recursive-CRP (streaming) \citep{schaeffer_efcient_2021}.
\end{itemize}

\begin{figure}[b]
\centering
\includegraphics[width=0.27\textwidth,trim={0 0 8cm 0},clip=True]{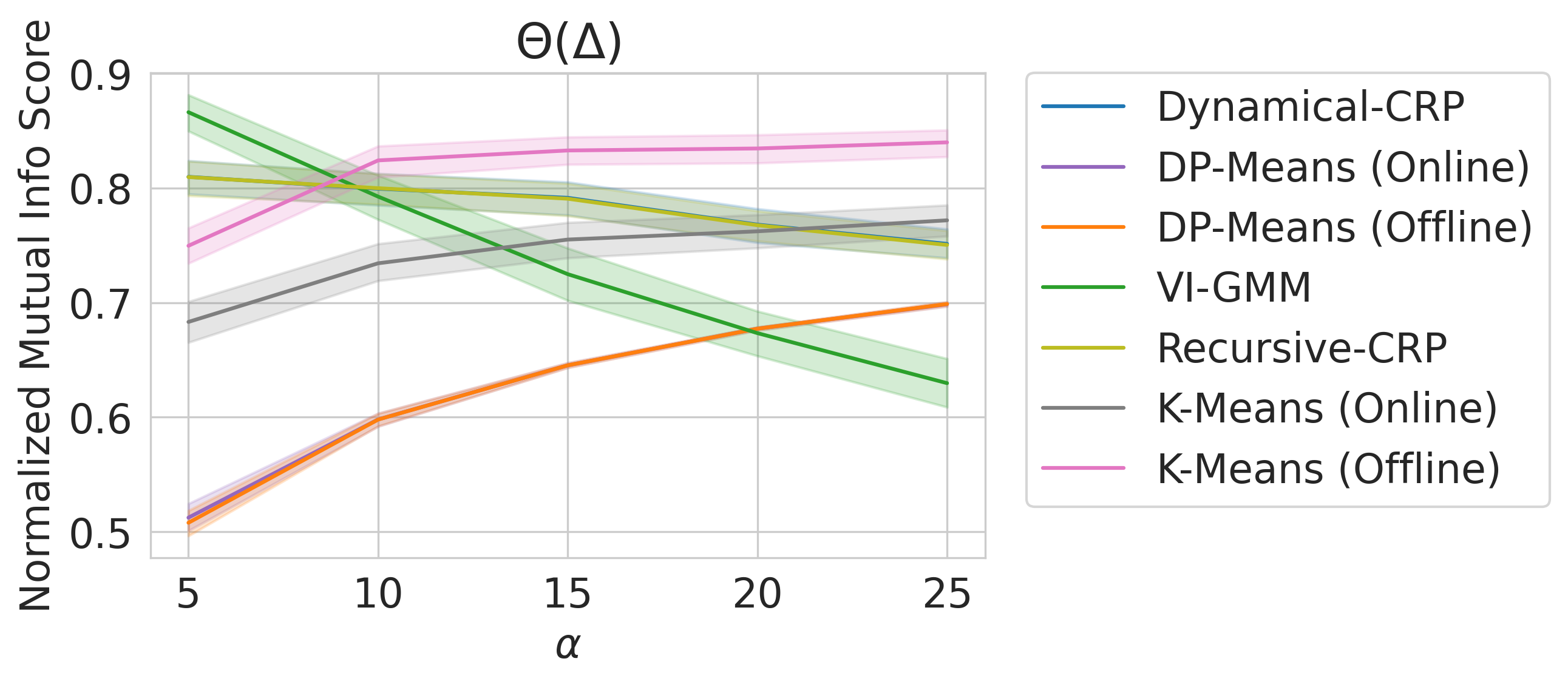}%
\includegraphics[width=0.27\textwidth,trim={0 0 8cm 0},clip=True]{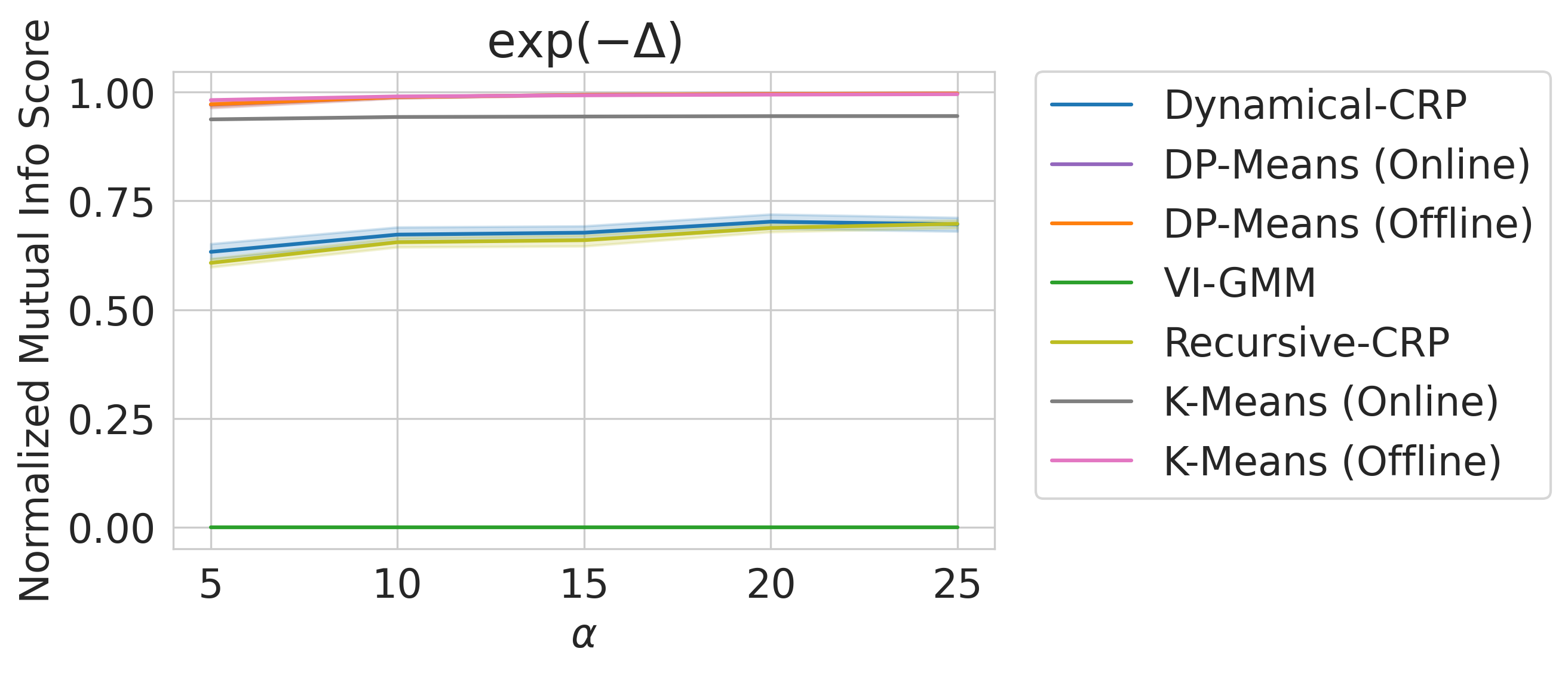}%
\includegraphics[width=0.21\textwidth, trim={0 0 8cm 0},clip=True]{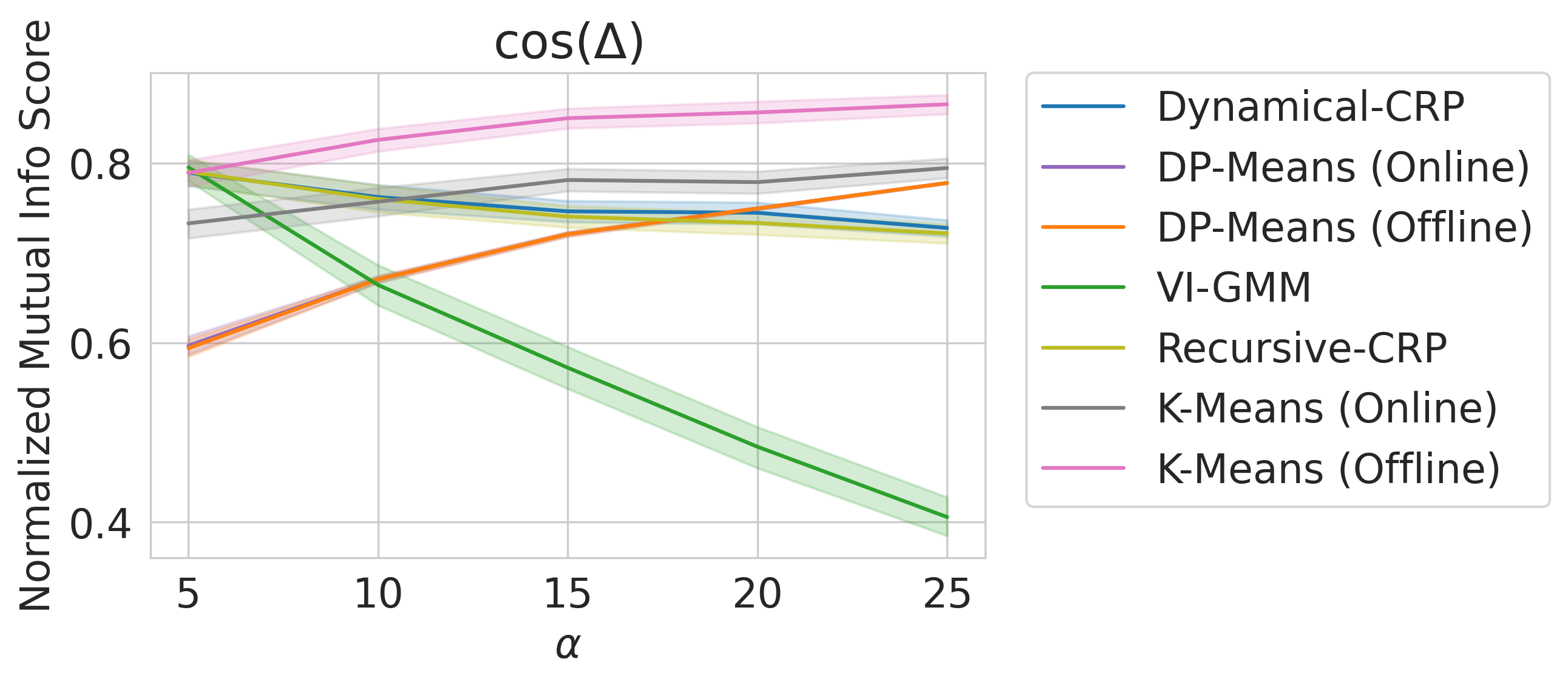}%
\includegraphics[width=0.45\textwidth]{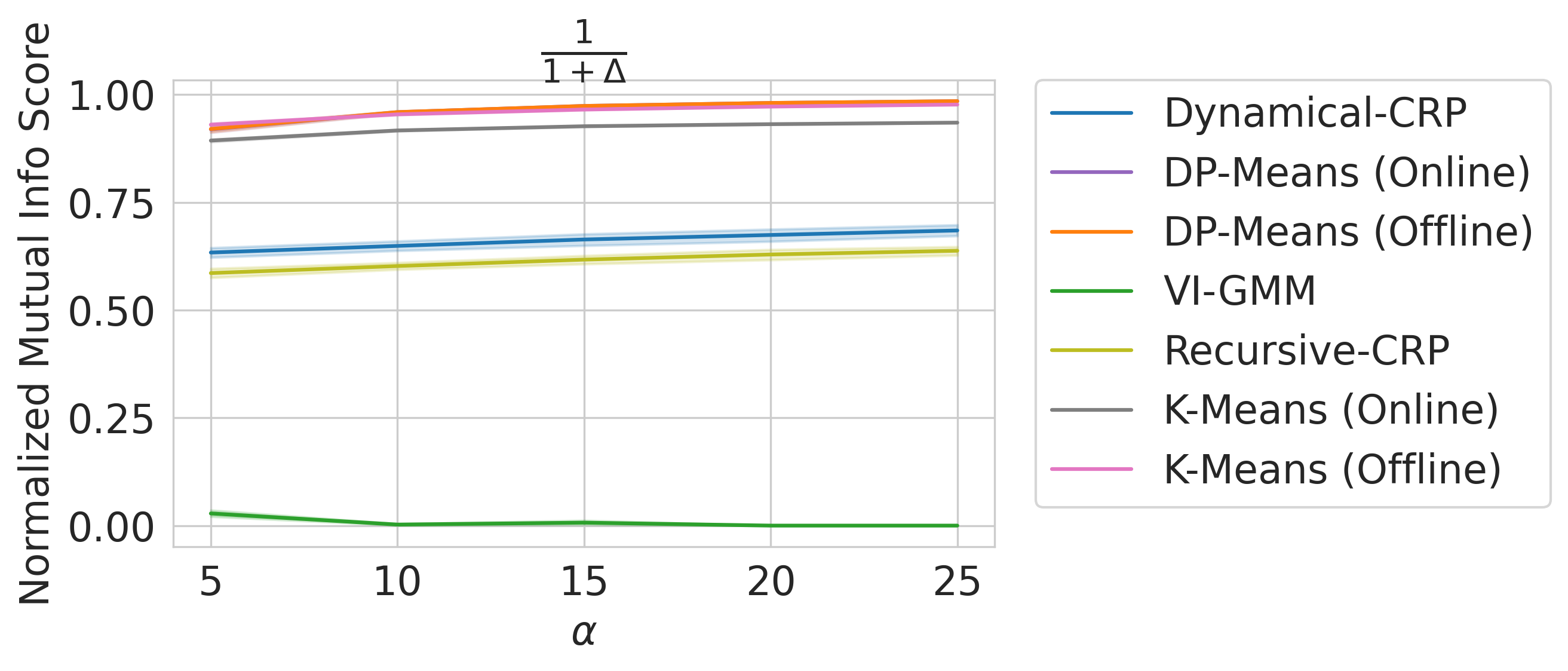}
\caption{\textbf{Normalized mutual information between true cluster assignments and inferred cluster assignments in Gaussian Mixture Models under 4 different dynamics.}}
\label{fig:01_mog_normalized_mutual_info_by_alpha}
\end{figure}

The baseline algorithms are all designed for stationary data, so we started our comparison with stationary dynamics, i.e. $\ell(\Tilde{N}) = \partial_t \Tilde{N}(t)$, but we also considered other dynamics (exponential, oscillatory, hyperbolic). We measured the performance of each algorithm by the (normalized) mutual information between the inferred cluster assignments and the true cluster assignments, implemented in scikit-learn \citep{pedregosa_scikit-learn_2011}. On stationary data, we found that Dynamical CRP was competitive on stationary data (Fig \ref{fig:01_mog_normalized_mutual_info_by_alpha}A), but excelled on non-stationary data (Fig \ref{fig:01_mog_normalized_mutual_info_by_alpha}BCD). 

\begin{figure}[t]
\centering
\includegraphics[width=0.21\textwidth,trim={0 0 8cm 0},clip=True]{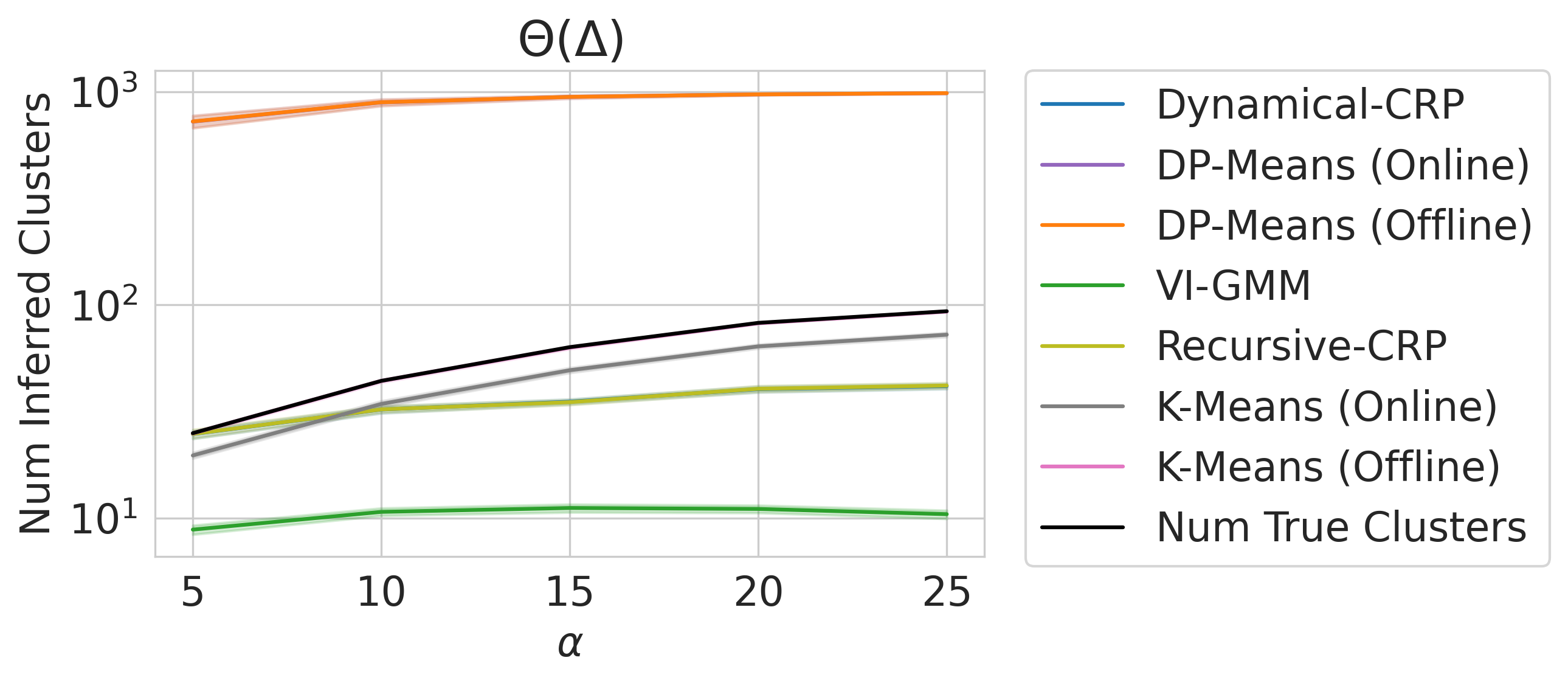}%
\includegraphics[width=0.21\textwidth,trim={0 0 8cm 0},clip=True]{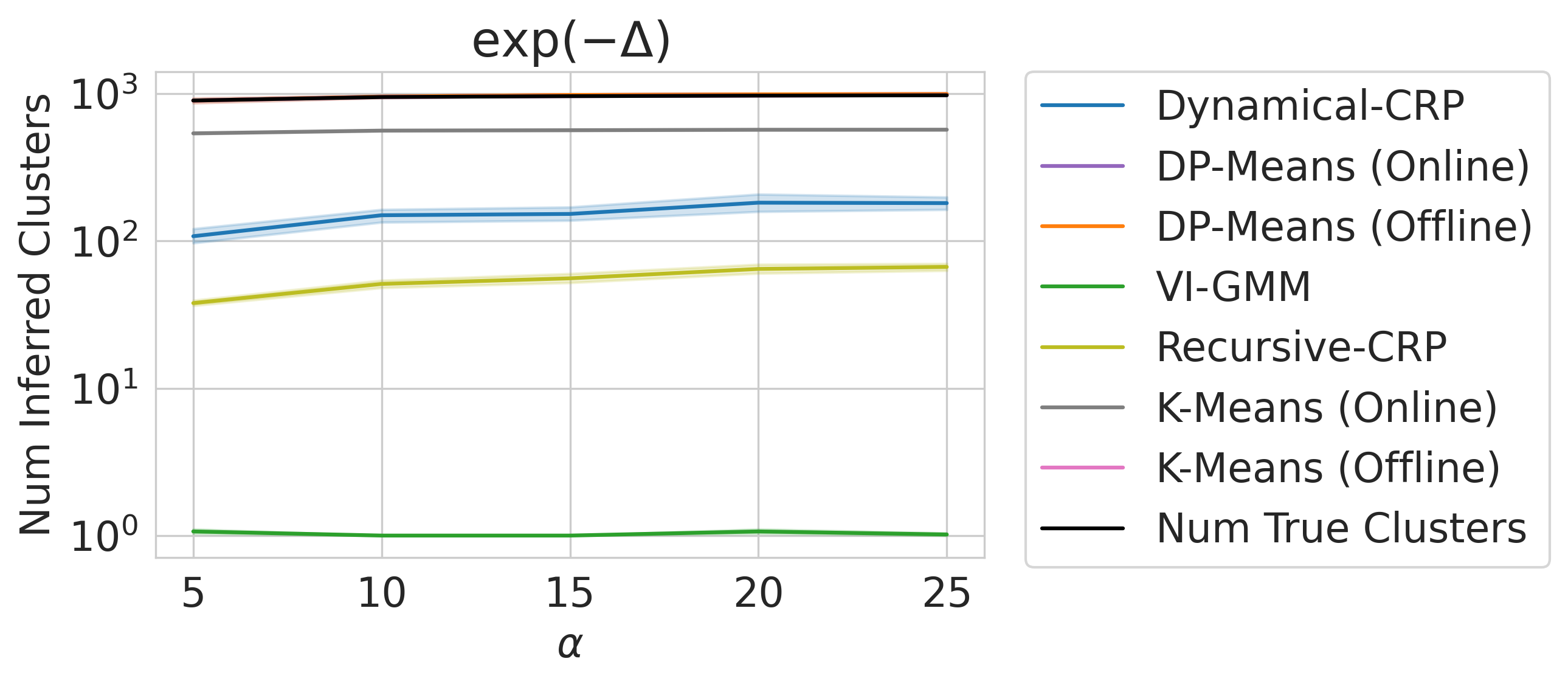}%
\includegraphics[width=0.21\textwidth, trim={0 0 8cm 0},clip=True]{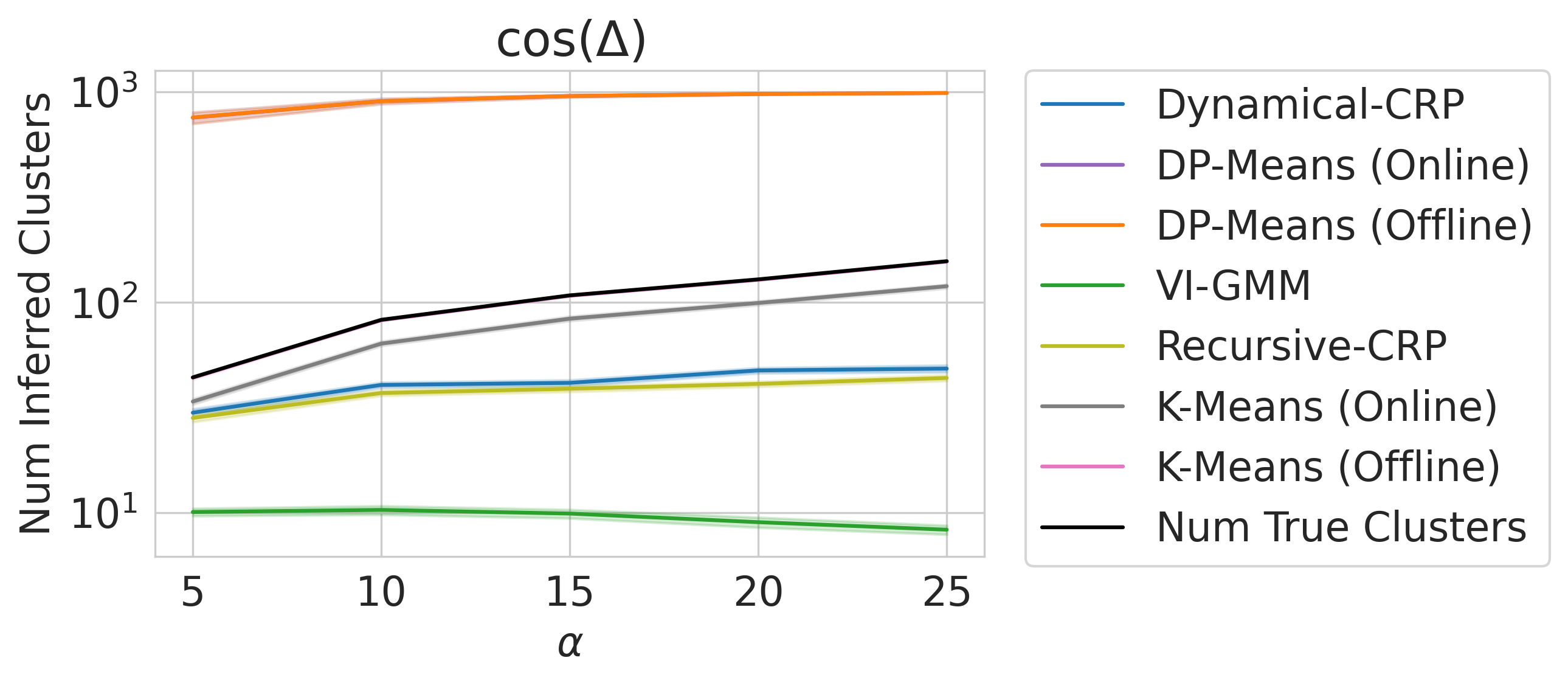}%
\includegraphics[width=0.34\textwidth]{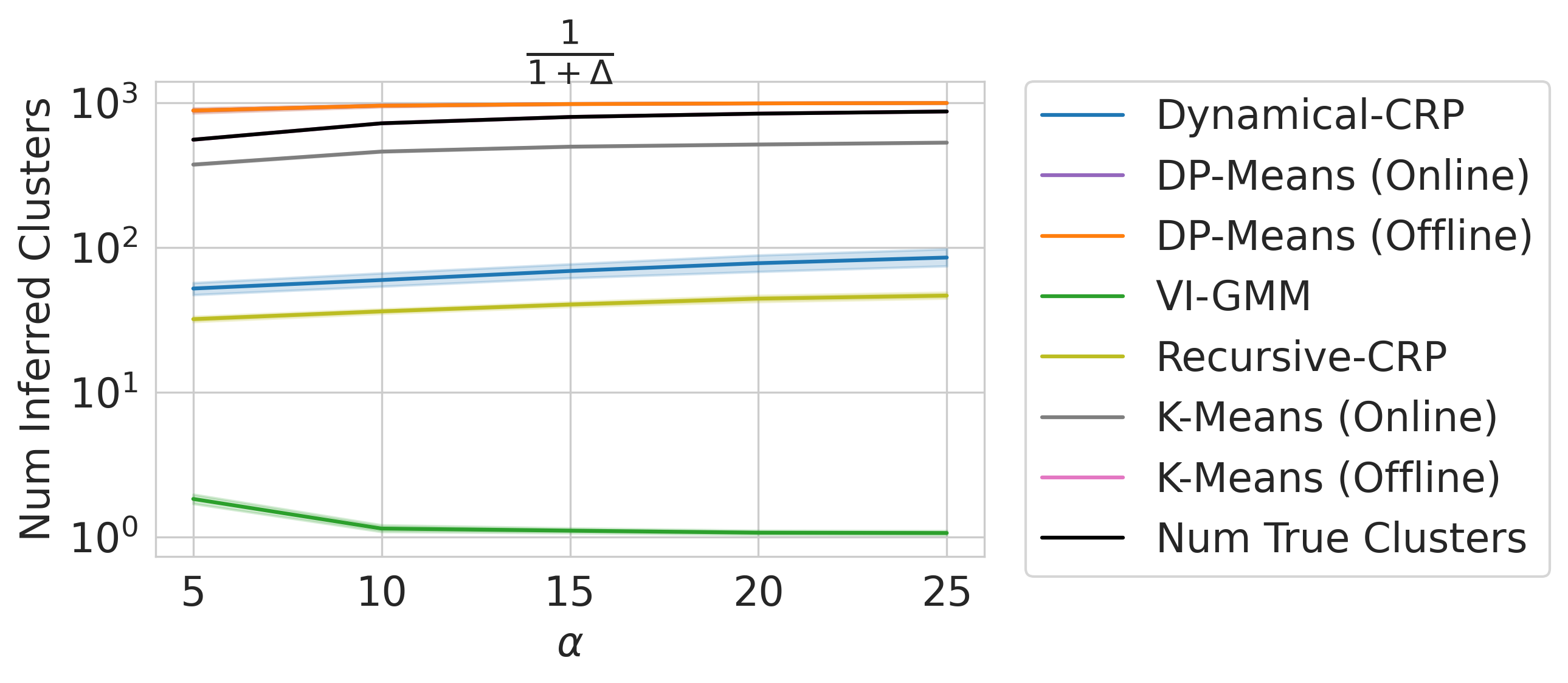}
\caption{\textbf{Dynamical CRP recovers close to the correct number of clusters under 4 different dynamics.}}
\label{fig:01_mog_num_inferred_clusters}
\end{figure}

\begin{figure}[t]
\centering
\includegraphics[width=0.24\textwidth]{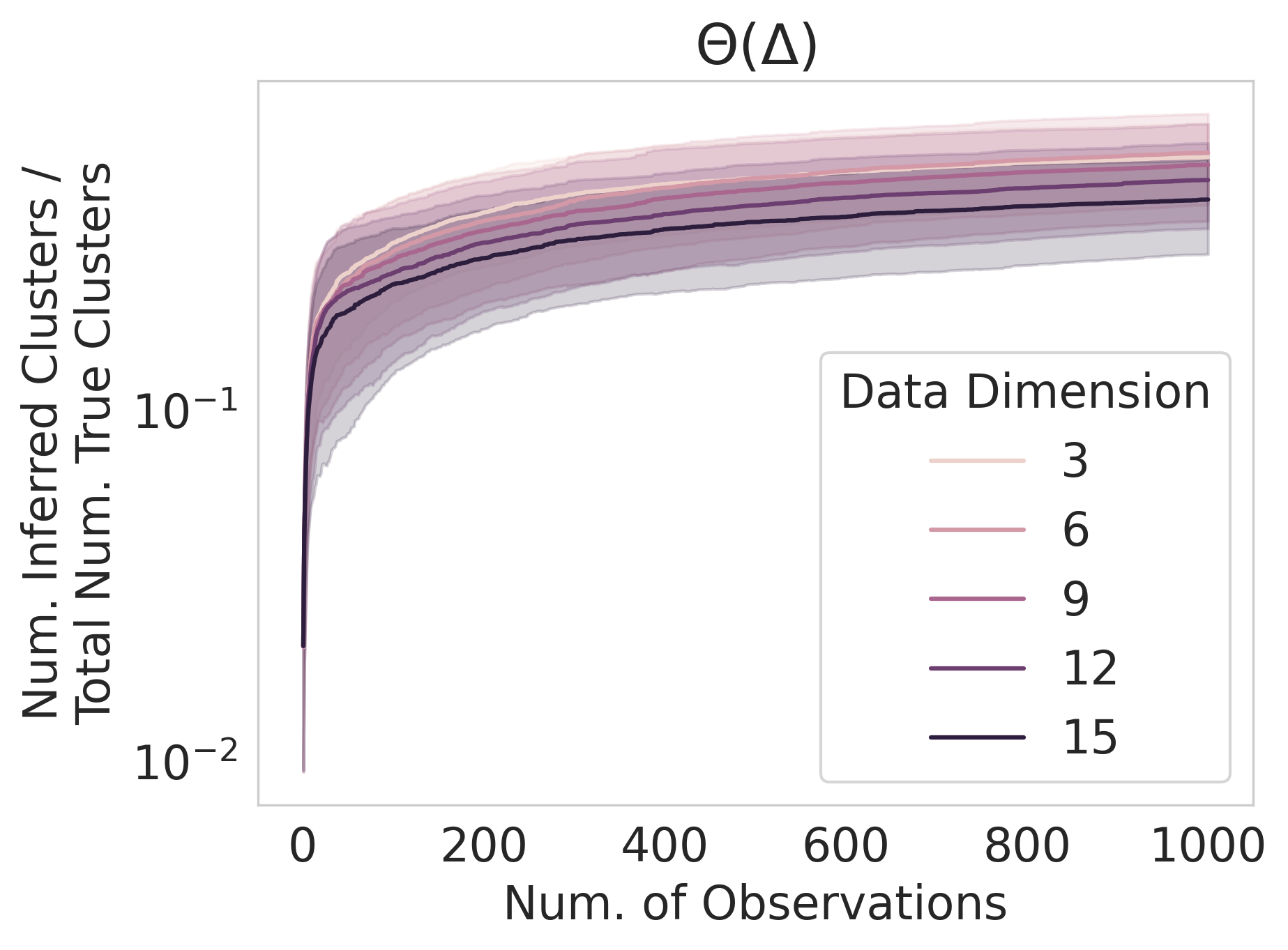}%
\includegraphics[width=0.24\textwidth]{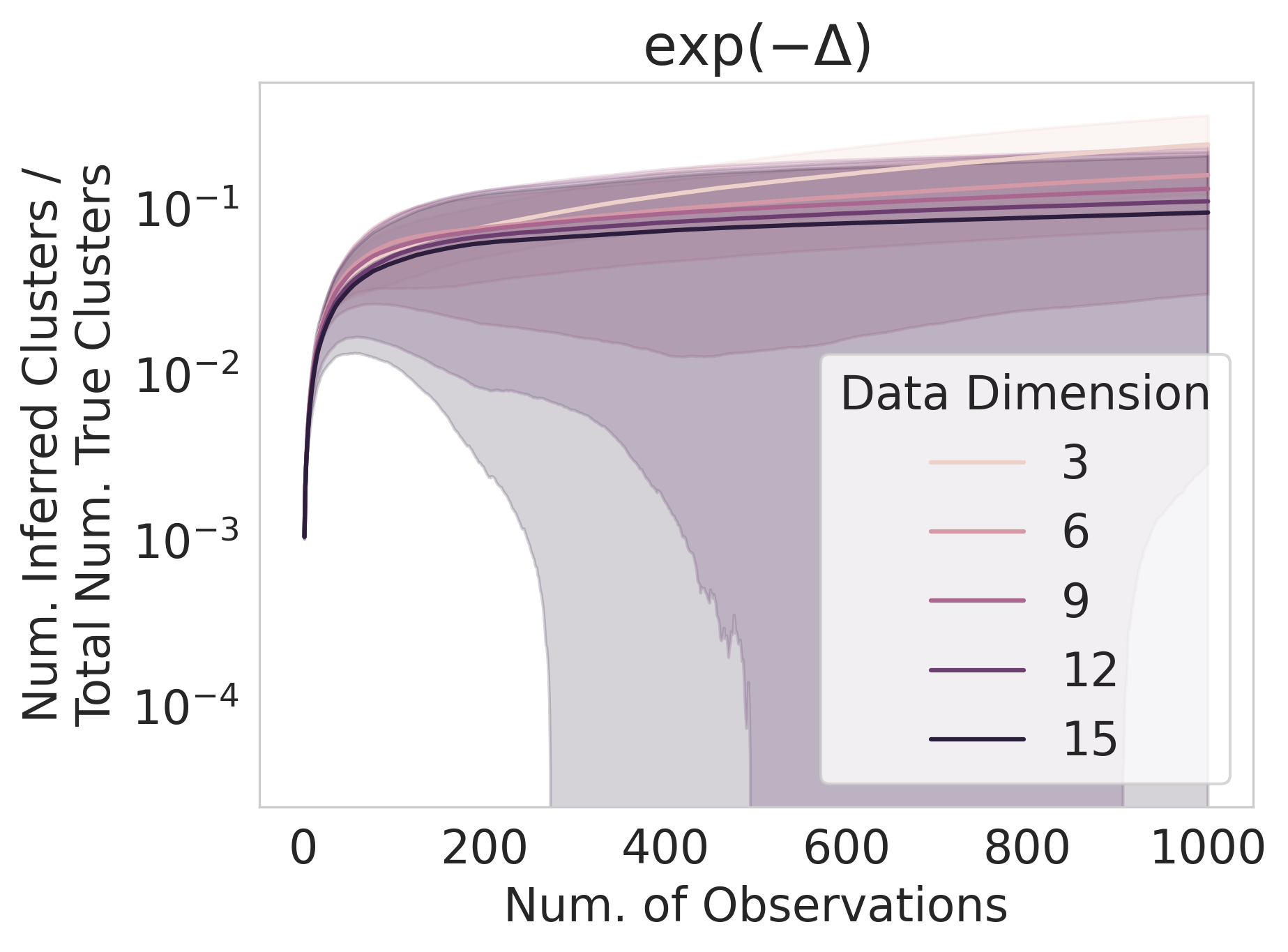}%
\includegraphics[width=0.24\textwidth]{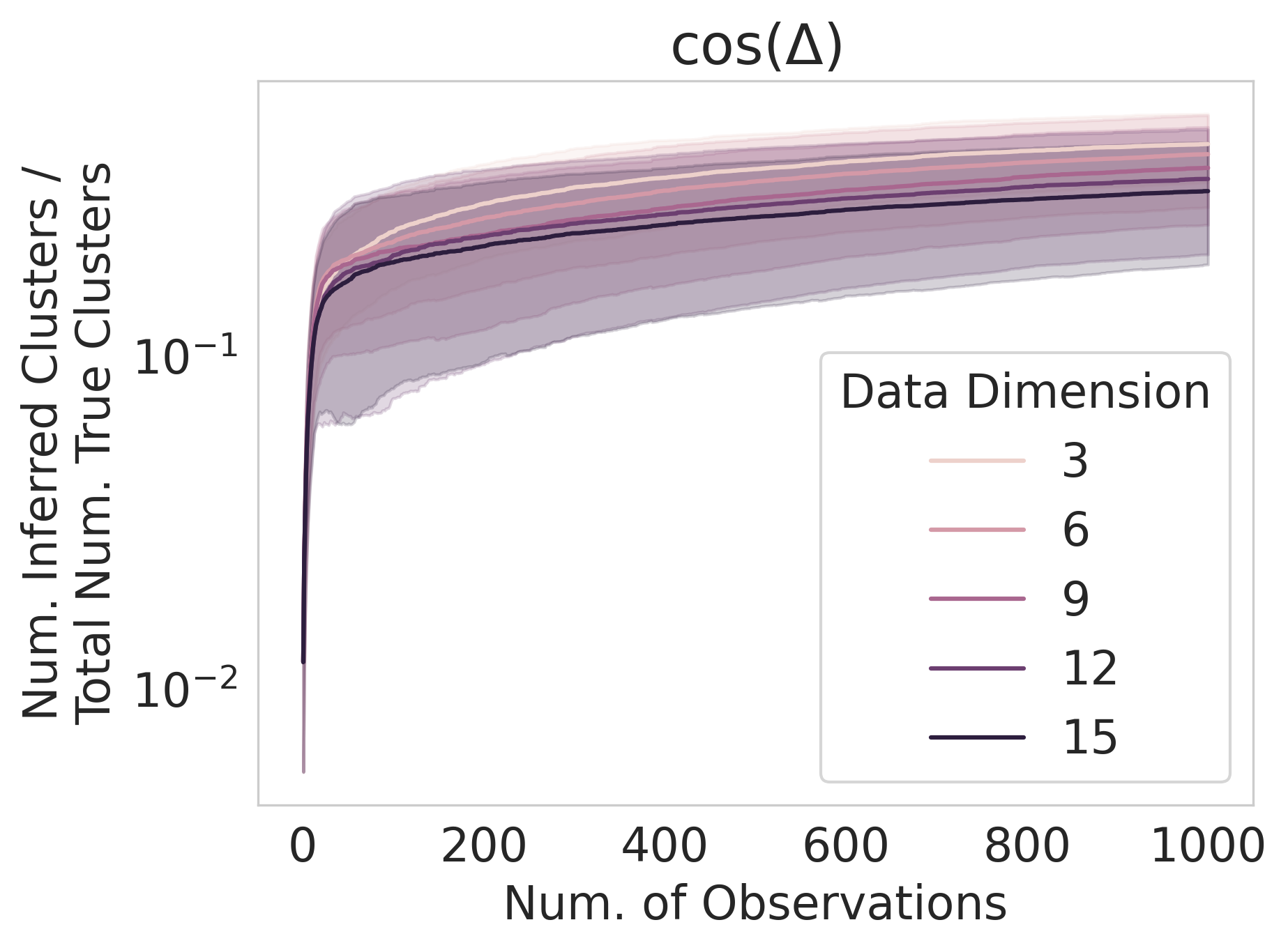}%
\includegraphics[width=0.24\textwidth]{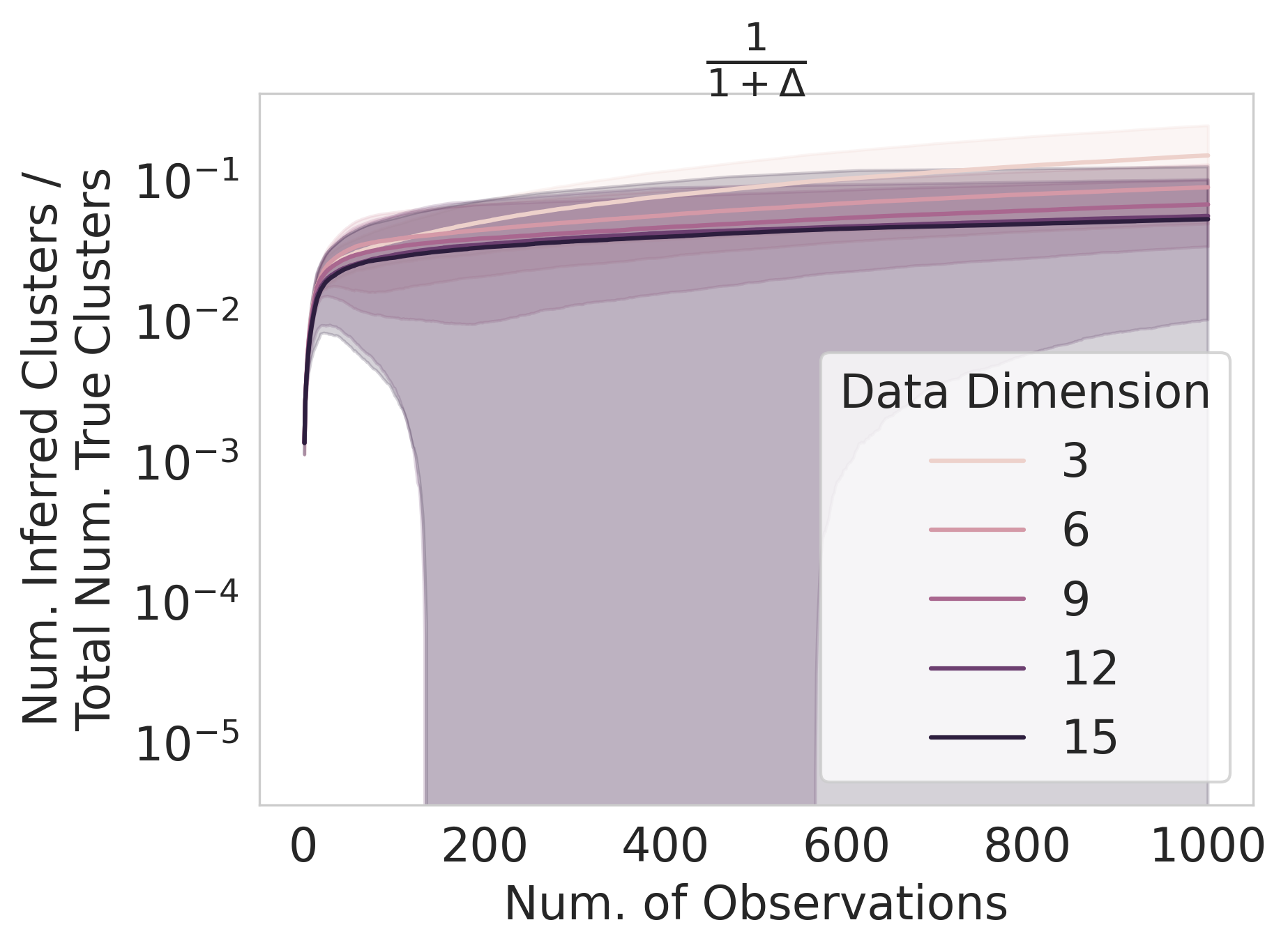}
\caption{\textbf{Dynamical CRP creates clusters over time, as necessitated by incoming data.}}
\label{fig:01_mog_frac_clusters_by_obs}
\end{figure}

We additionally plotted the number of clusters inferred by each algorithm. We found that D-CRP was often well within the correct order of magnitude, growing appropriately with the concentration hyperparameter $\alpha$ (Fig. \ref{fig:01_mog_num_inferred_clusters}). To explore how clusters are created as observations are received, we visualized when the Dynamical CRP creates clusters by plotting the ratio of the number of inferred clusters to the total number of true clusters as a function of the number of observations, dividing by the total number of true clusters in that set of observations. We found that Dynamical CRP creates clusters over time, as necessitated by the data (Fig. \ref{fig:01_mog_frac_clusters_by_obs}).

\begin{figure}[h]
\centering
\includegraphics[width=0.21\textwidth,trim={0 0 8cm 0},clip=True]{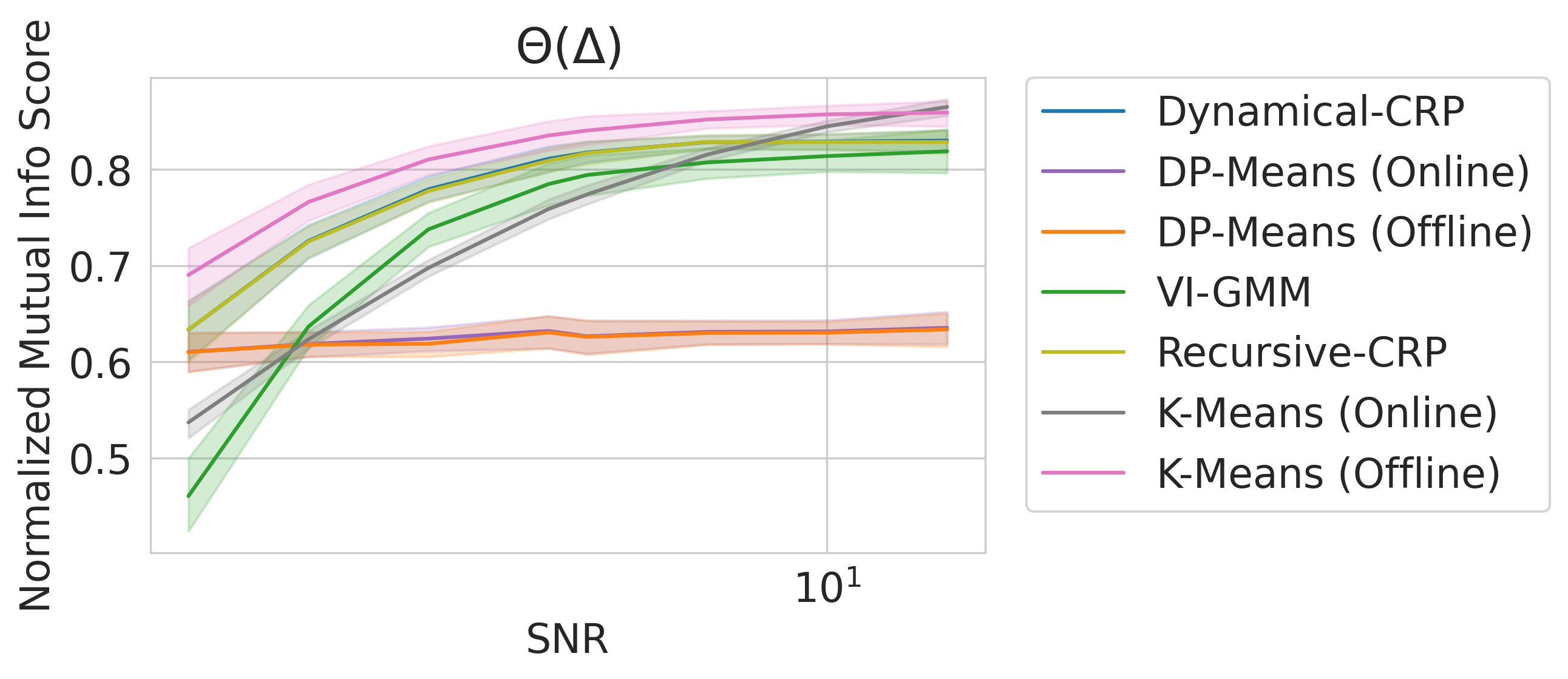}%
\includegraphics[width=0.21\textwidth,trim={0 0 8cm 0},clip=True]{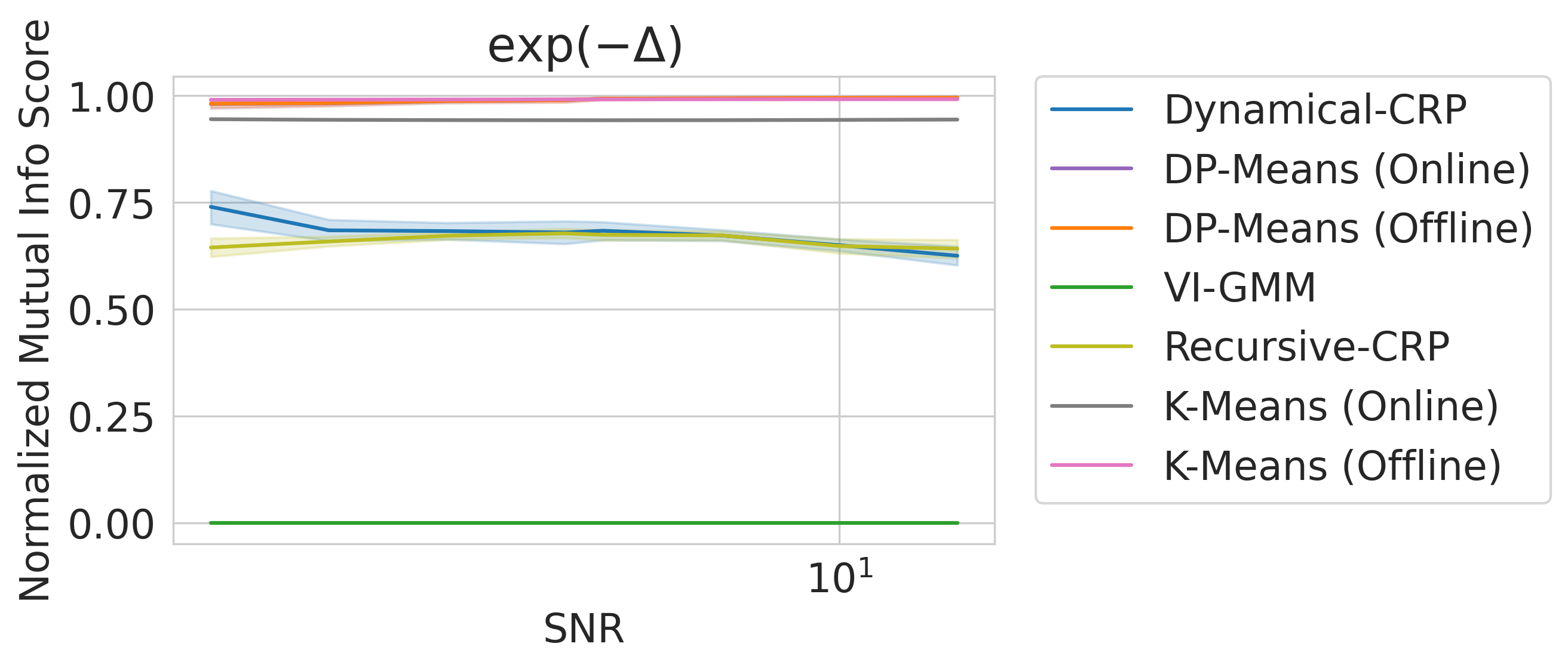}%
\includegraphics[width=0.21\textwidth,trim={0 0 8cm 0},clip=True]{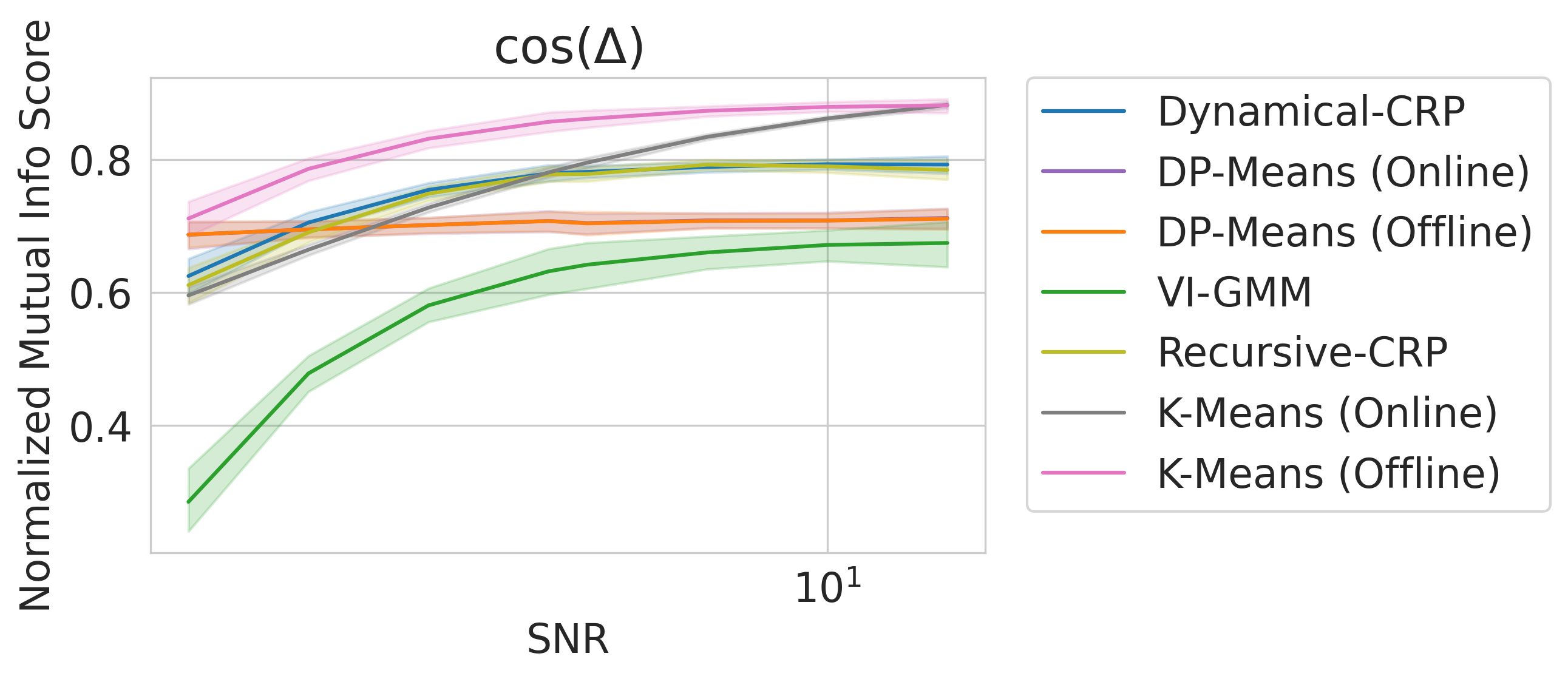}%
\includegraphics[width=0.34\textwidth]{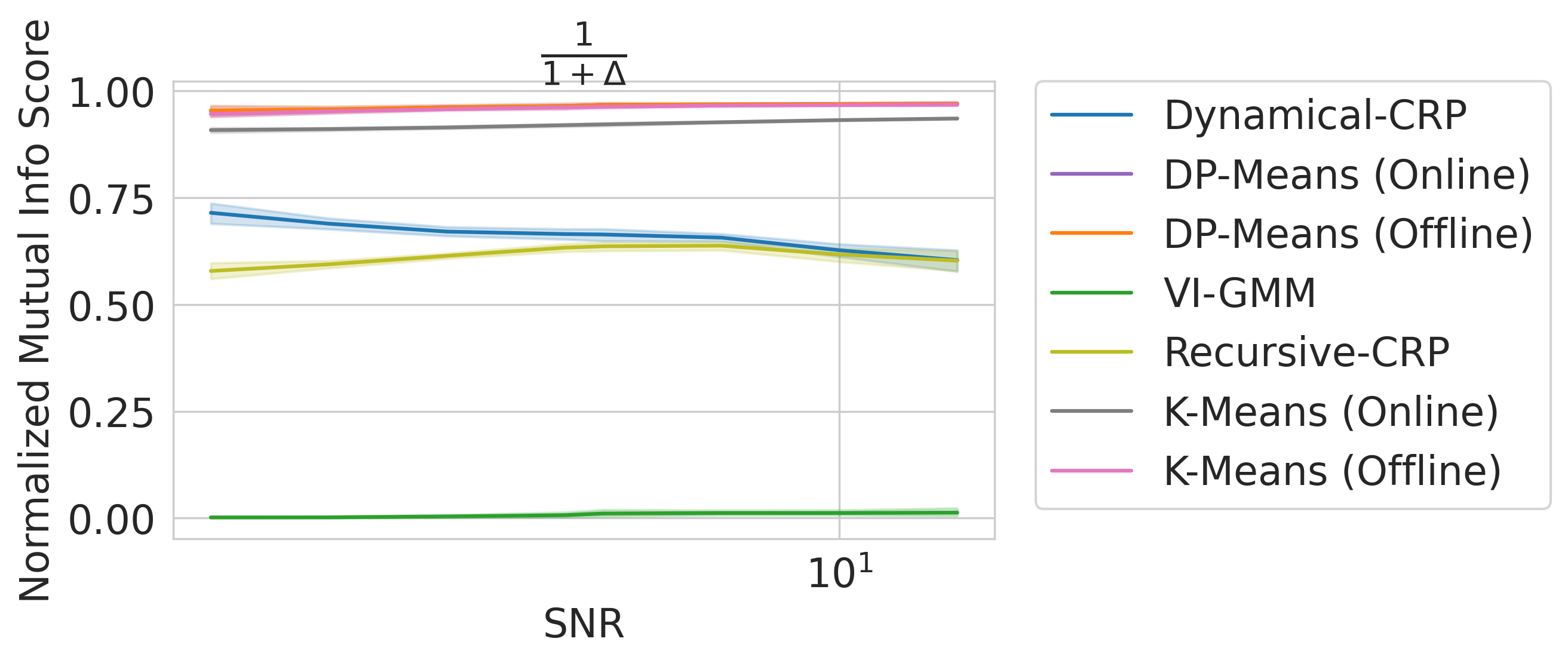}
\caption{\textbf{Dynamical CRP displays better performance when the data has a higher signal-to-noise ratio.}}
\label{fig:01_mog_normalized_mutual_info_by_snr}
\end{figure}

We also investigated how Dynamical CRP performs under different signal-to-noise (defined as the ratio of means covariance prefactor $\rho$ to likelihood covariance prefactor $\sigma$) regimes; interestingly, we found that while some algorithms do not display better performance with increasing signal-to-noise, Dynamical CRP does display better performance with higher SNR (Fig. \ref{fig:01_mog_normalized_mutual_info_by_snr}).

\subsection{Synthetic Mixture of von Mises-Fisher}

\begin{figure}[h]
\centering
\includegraphics[width=0.21\textwidth,trim={0 0 8cm 0},clip=True]{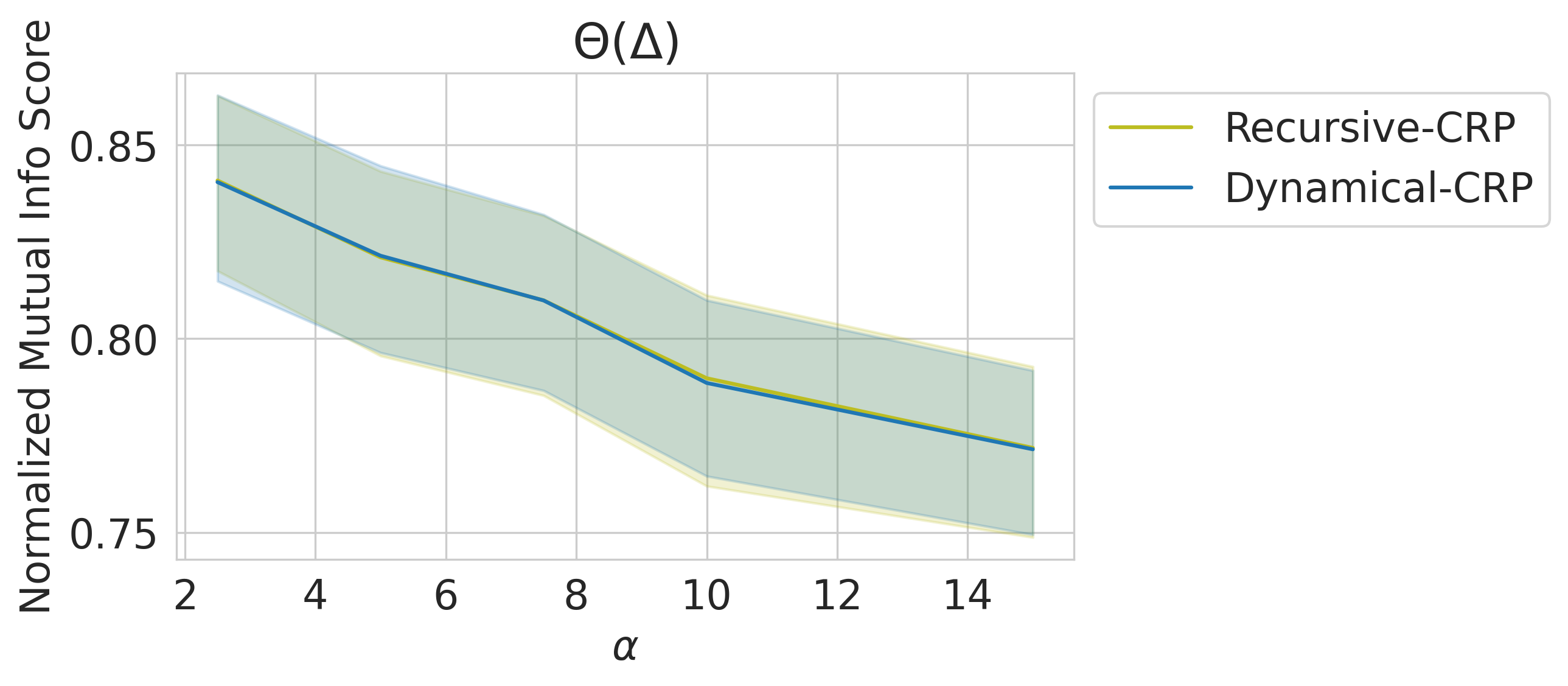}%
\includegraphics[width=0.21\textwidth,trim={0 0 8cm 0},clip=True]{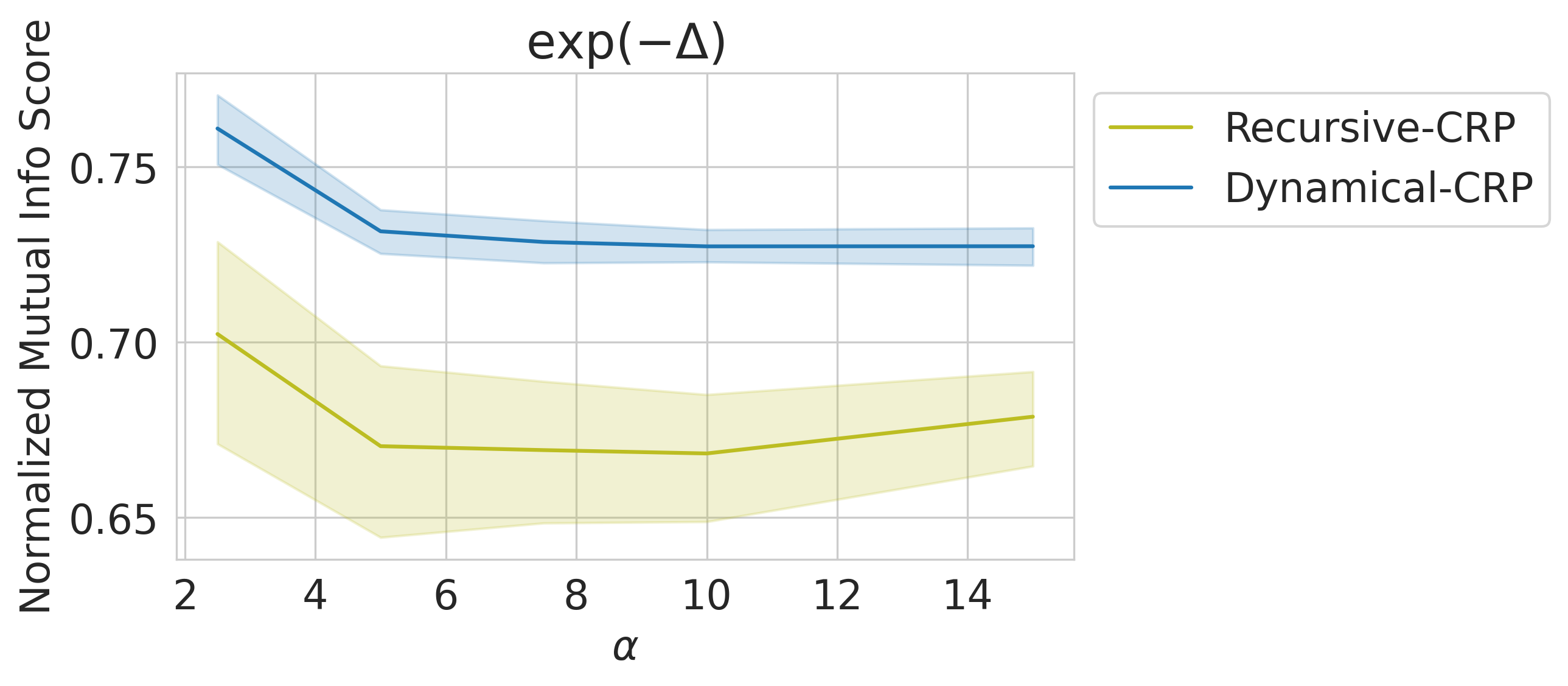}%
\includegraphics[width=0.21\textwidth, trim={0 0 8cm 0},clip=True]{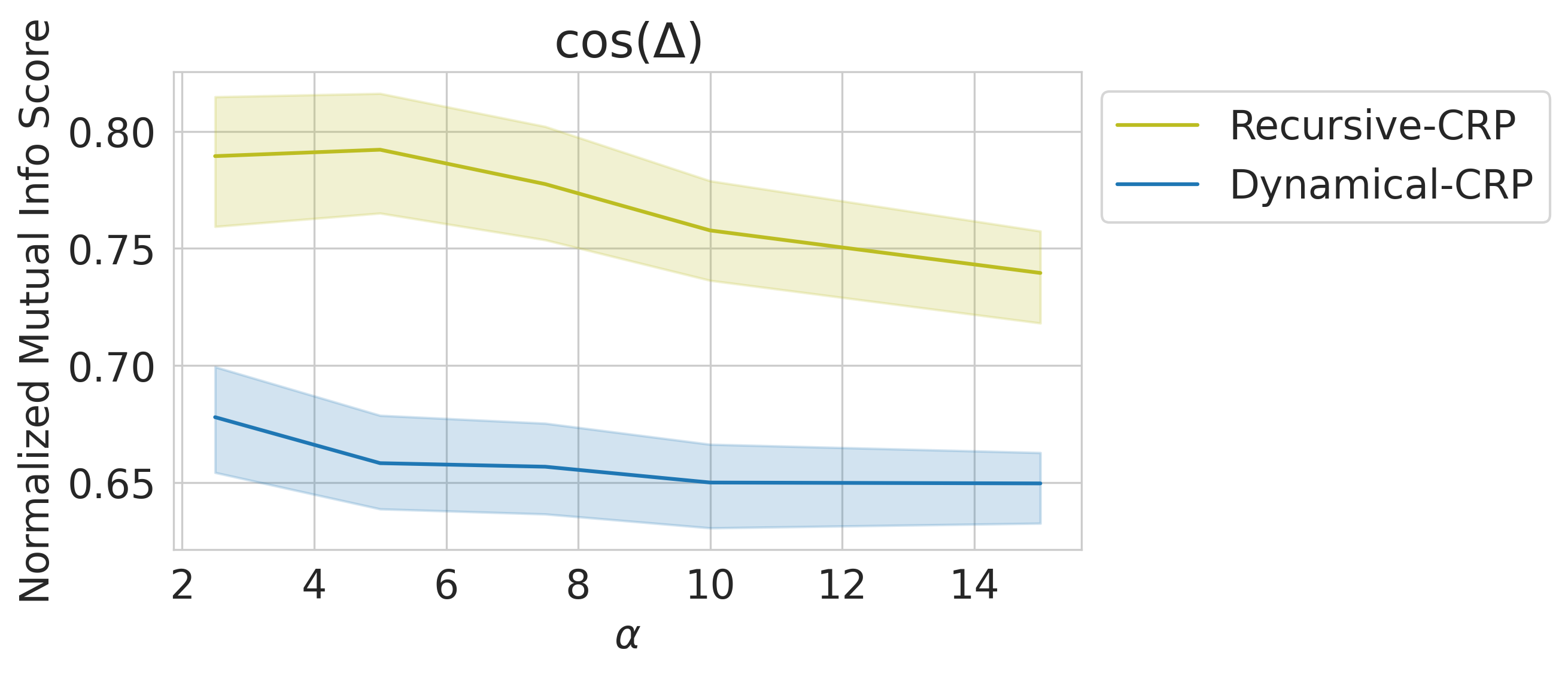}%
\includegraphics[width=0.33\textwidth]{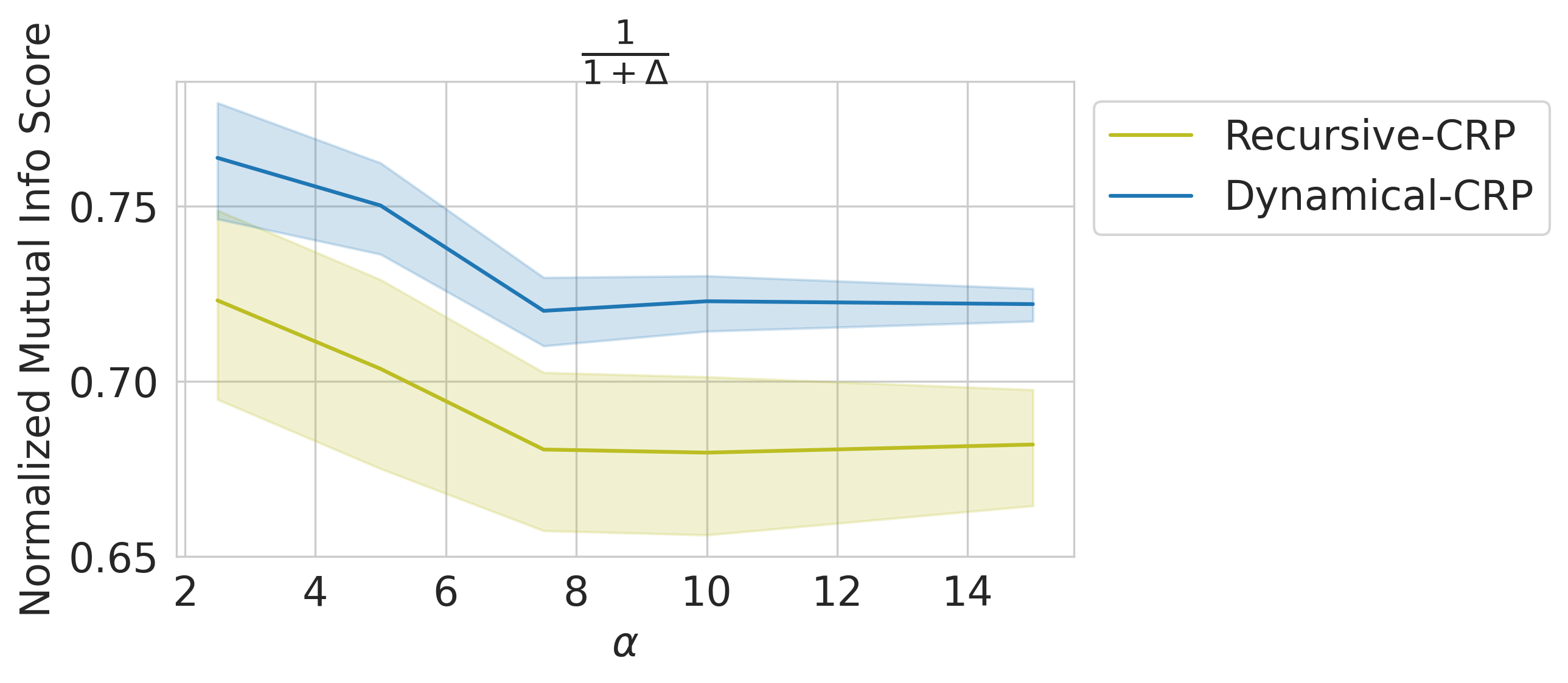}
\caption{\textbf{Normalized mutual information between true cluster assignments and inferred cluster assignments in von Mises-Fisher Mixture Models under 4 different dynamics.}}
\label{fig:01_movmf_normalized_mutual_info_by_alpha}
\end{figure}

\begin{figure}[h]
\centering
\includegraphics[width=0.21\textwidth,trim={0 0 8cm 0},clip=True]{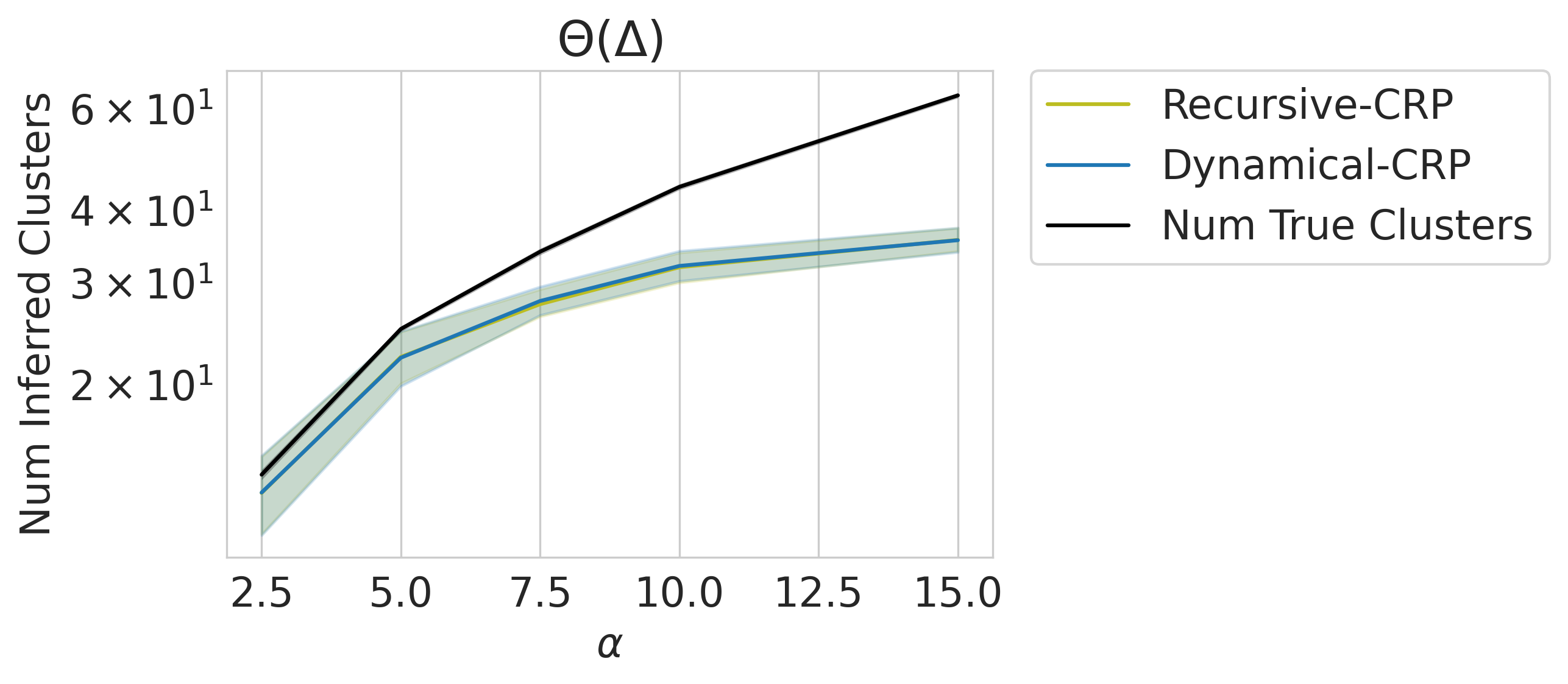}%
\includegraphics[width=0.21\textwidth,trim={0 0 8cm 0},clip=True]{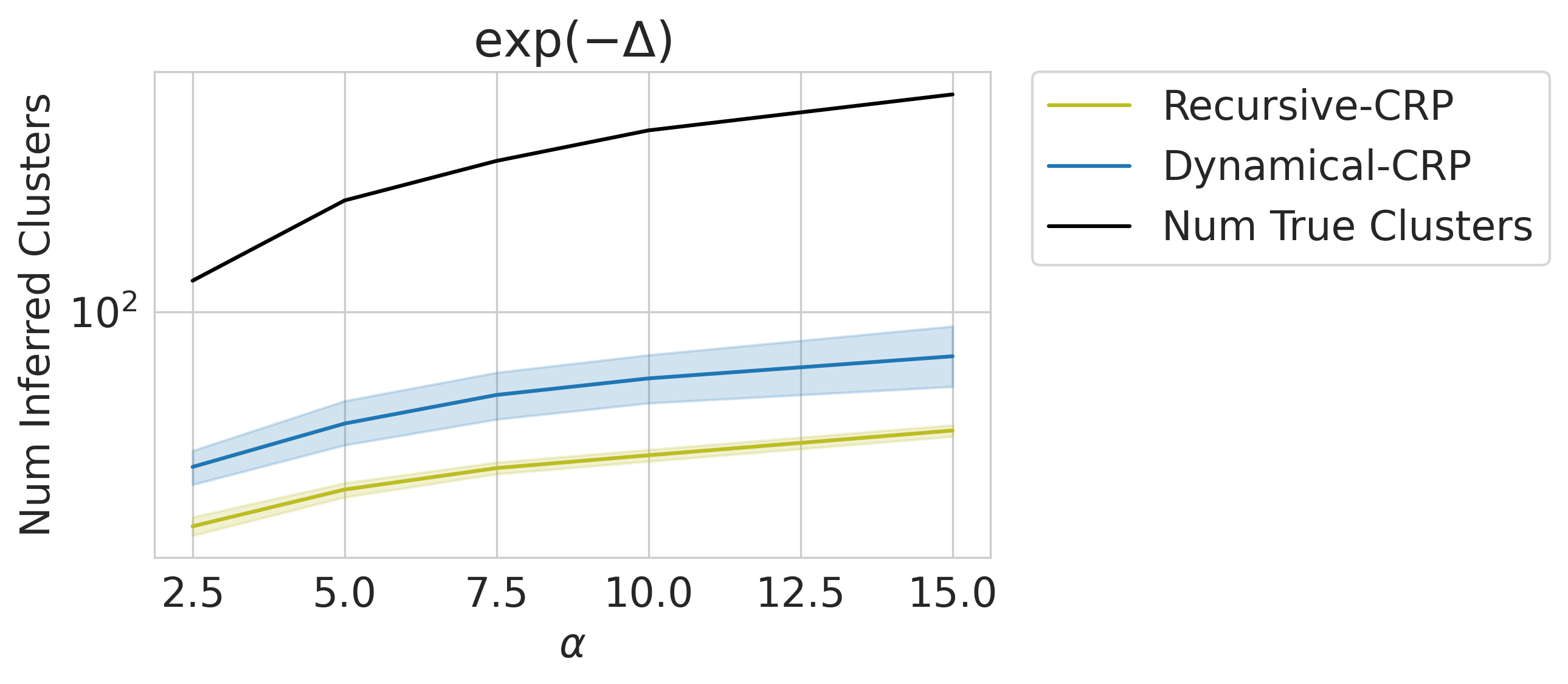}%
\includegraphics[width=0.21\textwidth, trim={0 0 8cm 0},clip=True]{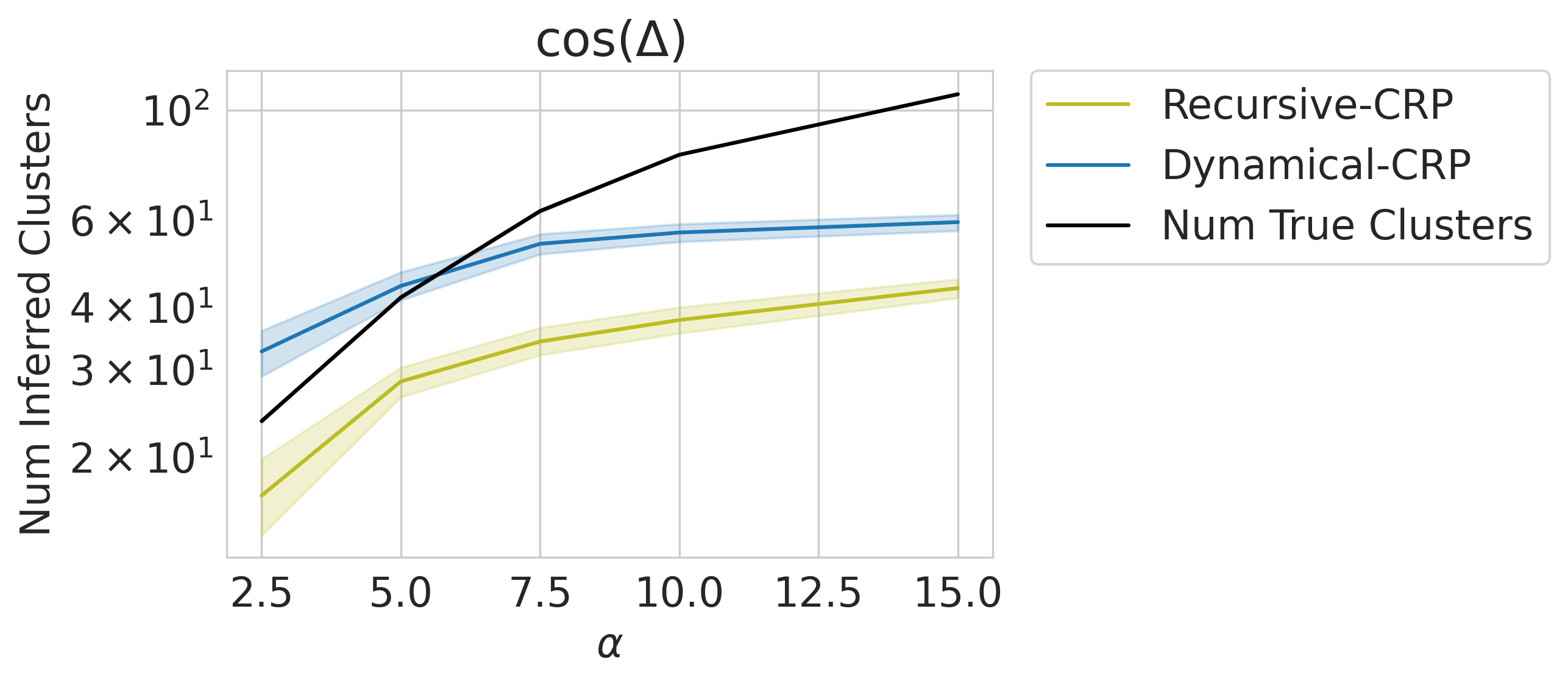}%
\includegraphics[width=0.34\textwidth]{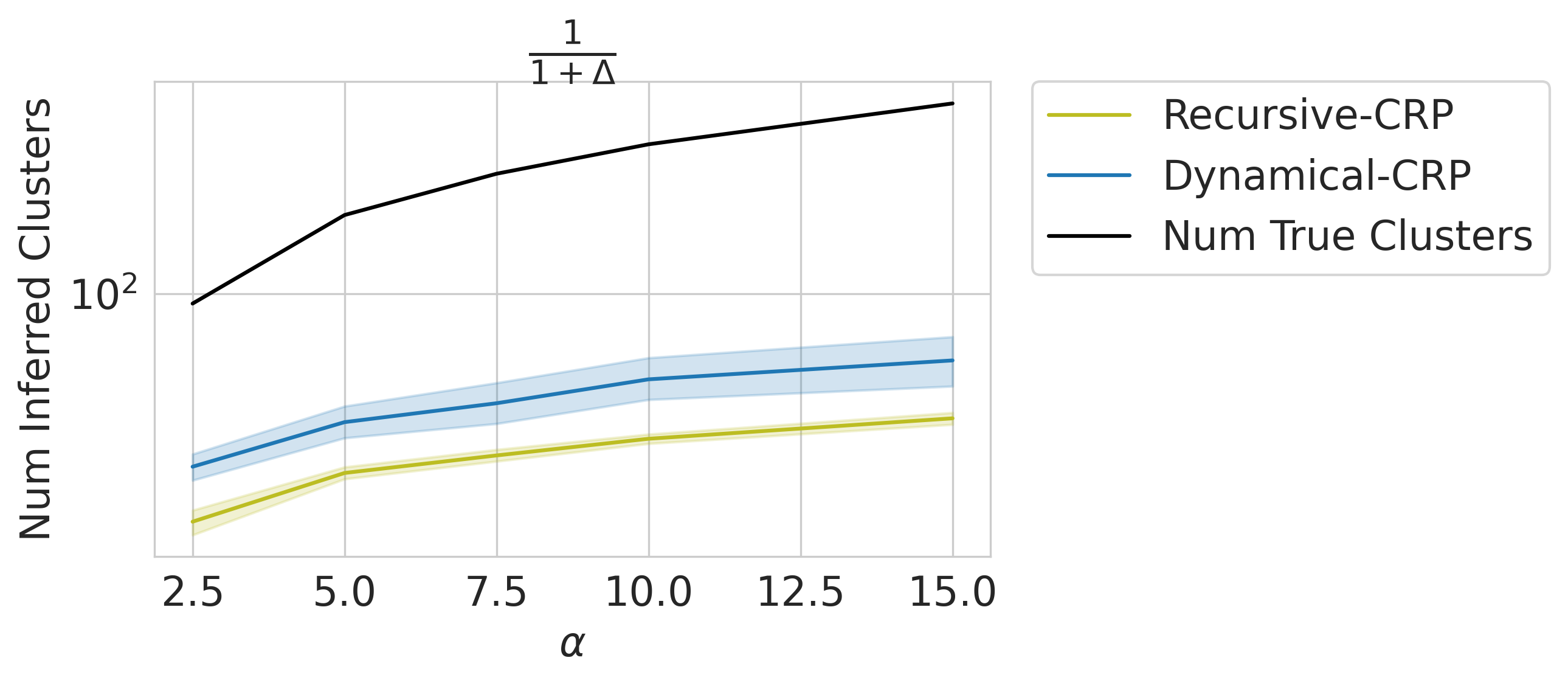}
\caption{\textbf{Dynamical CRP recovers close to the correct number of clusters under 4 different dynamics.}}
\label{fig:01_movmf_num_inferred_clusters}
\end{figure}

To demonstrate that the Dynamical CRP is not limited to Gaussian mixture models in Euclidean space, we turned to von Mises-Fisher mixture models on the surface of hyperspheres. We made this particular choice because one future line of work we are excited by involves combining deep learning with Bayesian nonparametrics for lifelong learning, and recent advances in self-supervised representation learning constrain deep neural network representations to the surface of hyperspheres \citep{chen_simple_2020, grill_bootstrap_2020, caron_unsupervised_2021}. As with the mixture of Gaussians, we generated datasets of 1000 observations from the generative model, with a uniform prior on the cluster directions $p(\phi) = \mathcal{VMF}(\kappa=0)$, by sweeping over dynamics, alpha, signal-to-noise ratios, and ambient dimension to construct 3600 datasets total. Most previous baselines were designed for Gaussian likelihoods, meaning only the Recursive-CRP could be used. We again plotted the number of clusters inferred by each algorithm. We found that D-CRP often outperformed R-CRP (Fig. \ref{fig:01_movmf_normalized_mutual_info_by_alpha}) and was often well within the correct order of magnitude, growing appropriately with the concentration hyperparameter $\alpha$ (Fig. \ref{fig:01_movmf_num_inferred_clusters}).

\subsection{Room Clustering for Simultaneous Localization and Mapping}

Clustering is useful not just in its own right, but also as a sub-task for other tasks. For instance, an agent may wish to cluster observations into states for use in planning or reinforcement learning. As a demonstration, we turn to the domain of simultaneous localization and mapping (SLAM) \citep{rosen_advances_2021}, in which an agent must both learn a map of its environment as well as its location within that environment. One common approach is to learn hierarchically \citep{fairfield_segmented_2010, klukas_fragmented_2022} by clustering sensory observations into rooms that can then be used to efficiently plan.

\begin{figure}[h]
\centering
\includegraphics[width=0.24\textwidth]{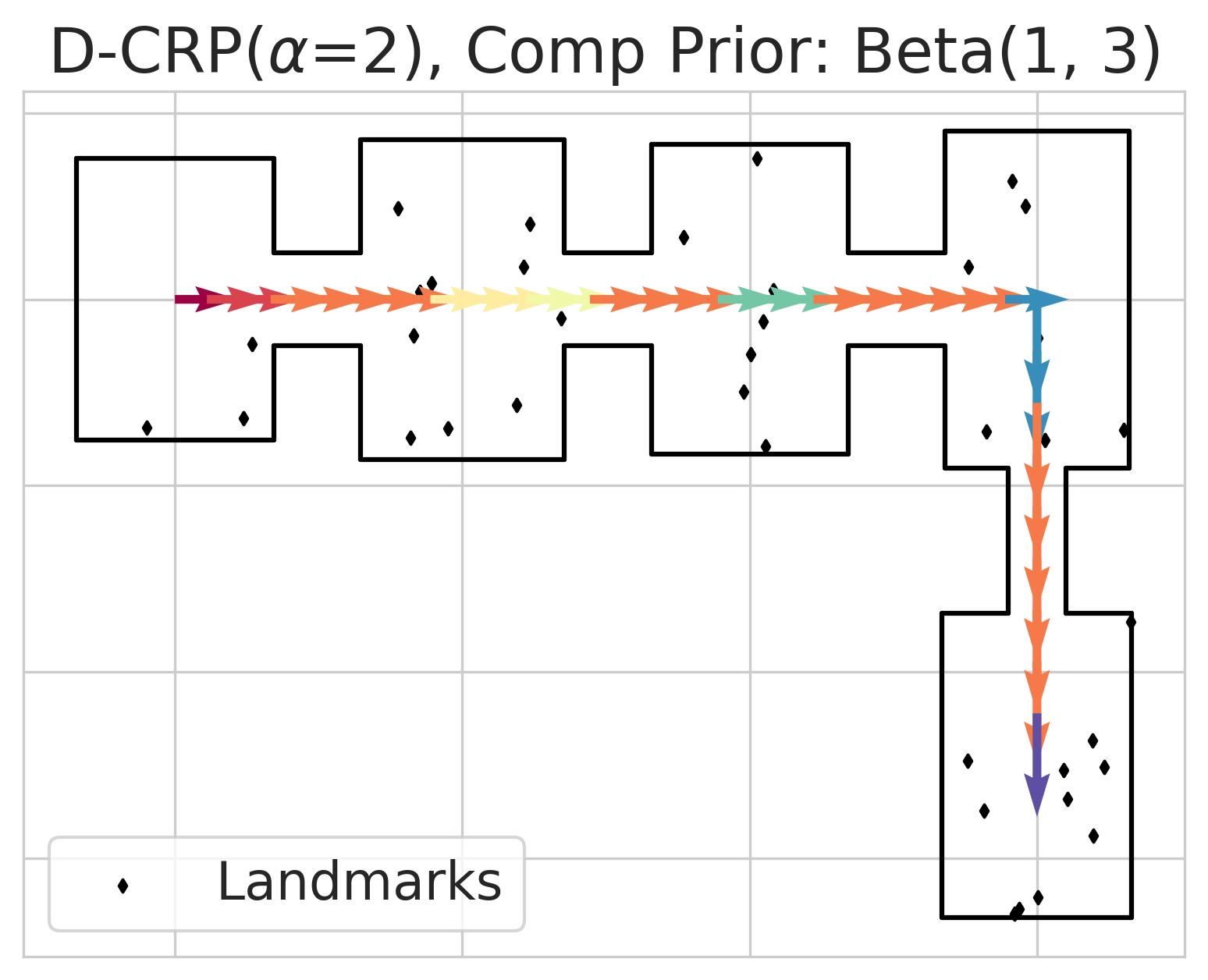}%
\includegraphics[width=0.24\textwidth]{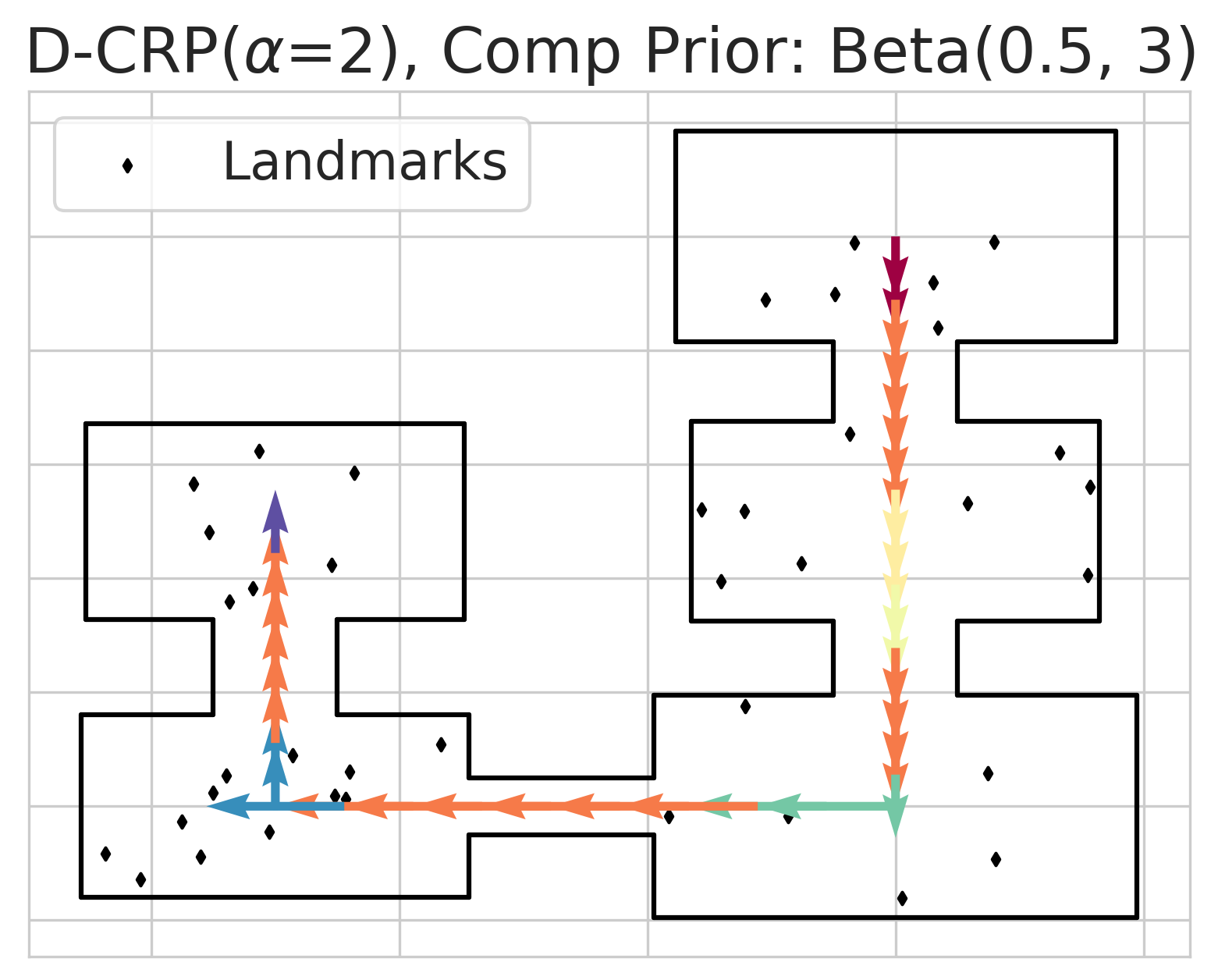}%
\includegraphics[width=0.24\textwidth]{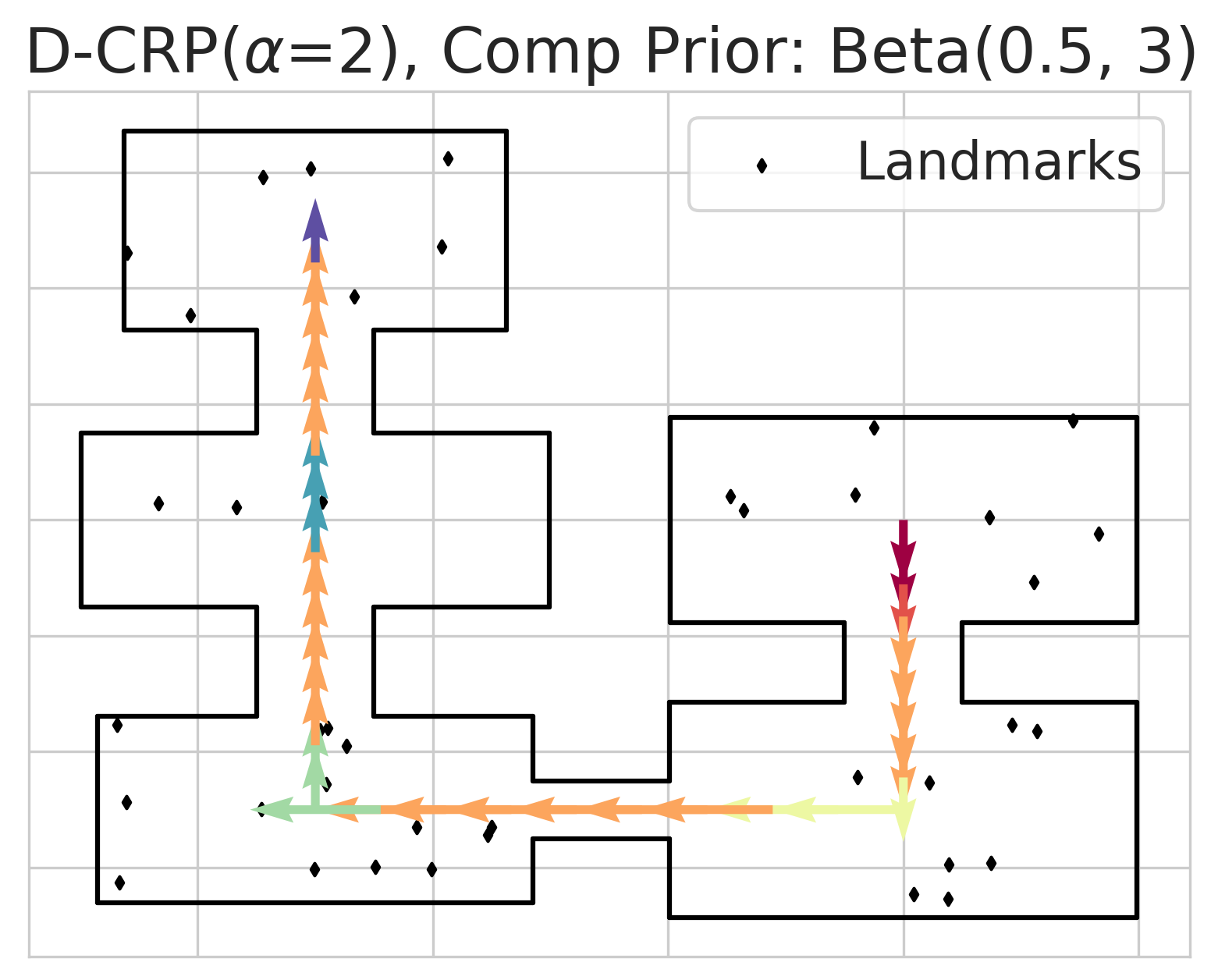}%
\includegraphics[width=0.24\textwidth]{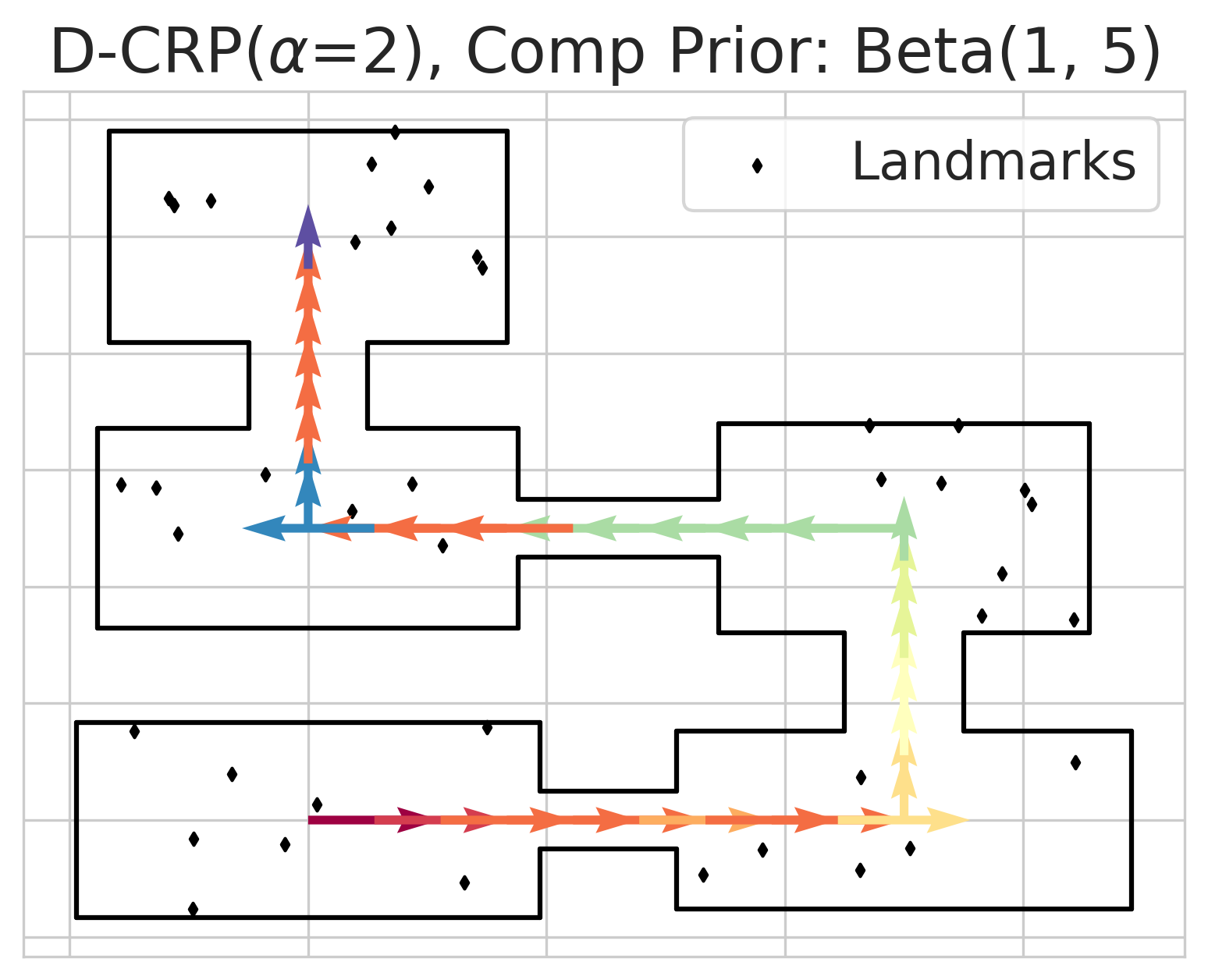}
\caption{\textbf{Clusters inferred by Dynamical CRP in a 2D spatial navigation task.} Each color in each environment represents a unique cluster, inferred from visible landmarks (black diamonds) encountered along a single trajectory. The Dynamical CRP aggregates visually-distinguishable rooms (various colors) into distinct clusters and visually-identical hallways into the same cluster (orange).}
\label{fig:yilun_nav_2d}
\end{figure}

We procedurally generated environments with multiple rooms, each containing a variable number of sensory observations (``landmarks" in the SLAM literature), and then applied D-CRP with exponential dynamics to a single trajectory through each novel environment. At each position along the trajectory, a landmark is either visible or not, determined by a limited range of view. We used a product-of-Bernoullis likelihood for expressing whether each landmark is visible from a given position; for simplicity, the likelihood does not take into account position or velocity. We found the D-CRP with exponentially decaying dynamics excels at this task. As D-CRP takes a single trajectory through each novel environment, D-CRP aggregates sensory landmarks (black diamonds) into visually-distinguishable unique clusters (rooms) and visually-indistinguishable non-unique clusters (hallways, orange) (Fig \ref{fig:yilun_nav_2d}).

The D-CRP excels at this task when equipped with exponentially decaying dynamics, because the dynamics impose an inductive bias that trajectories are temporally and spatially smooth, meaning the D-CRP has a strong prior towards allocating two sequential observations to the same cluster even if the two sensory observations differ significantly. These rooms could then serve as a high-level representation for hierarchical spatial navigation.

%% file: 05_discussion.tex
\section{Discussion}

In this paper, we attack unsupervised learning on streaming non-stationary data in the specific setting of mixture modeling. We propose a novel stochastic process that defines a non-exchangable distribution over partitions of a set, that we termed the Dynamical Chinese Restaurant Process. We show that the Dynamical CRP provides a bespoke non-stationary prior over cluster assignments and is amenable to an efficient streaming variational inference algorithm. We then demonstrate on both synthetic and real data, with Gaussian and non-Gaussian likelihoods, that the Dynamical CRP provides a powerful clustering algorithm for non-stationary streaming data.

%% file: 06_conclusion.tex


%% file: 07_appendix.tex
\section{Coordinate Ascent Variational Inference Parameter Updates}

This section contains derivations for Coordinate Ascent Variational Inference (CAVI). Specifically, the following derivations show how to update variational parameters in different with various likelihoods.

\subsection{Closed-Form Expression for Variational Parameters in the Exponential Family}

In the following three subsections, we use the following fact from \citet{beal_variational_2003, wainwright_graphical_2008}: if a distribution $p$ and its variational approximation $q$ are both in the exponential family, then the optimal variational parameters $\zeta_i$ that correspond to the variational distribution over variable $W_i$ are the solution to
\begin{equation}
    \log q(W_i; \zeta_i) = \mathbb{E}_{q(W_{-i})}[\log p(W, X | \theta)]
    \label{eq:exp_family_var_solution}
\end{equation}

This means that when optimizing the parameters for one variable, we can replace all other variables with their expectations under the variational distribution and then solve for that one variable's variational parameters.

\subsection{CAVI for Multivariate Gaussian Likelihood}

Mean field family:
\begin{align*}
    q(c_n, \{\phi \} | o_{\leq n}) \defeq q(c_n|o_{\leq n}; \{\pi_{nc}\})& \prod_{k=1}^{C_n} q(\phi_{nc}|o_{\leq n}; \mu_{nc}, \Sigma_{nc}) \\
    q(c_n|o_{\leq n}; \{\pi_{nc} \}) &\defeq Categorical(\pi_n)\\
    q(\phi_{nc}|o_{\leq n}; \mu_{nc}, \Sigma_{nc}) &\defeq \mathcal{N}(\mu_{nc}, \Sigma_{nc})
\end{align*}

where $\theta_n \defeq \{\pi_{nc}\}_k \cup \{\mu_{nc}\}_k \cup \{\Sigma_{nc}\}_k$ are our variational parameters for the $n$th observation. The mixture weights' parameters $\pi_n$ will be determined by solving the following:
\begin{align*}
    \log q(c_n|o_{\leq n}; \pi_{n}) &= \mathbb{E}_{q(\{\phi_{nc}\})}[\log p(o_n, c_n, \{\phi_{nc}\}|o_{<n})]
\end{align*}

The left-hand side (LHS) is:
\begin{align*}
    &\log q(c_n|o_{\leq n}; \pi_{n}) = \sum_k \mathbbm{I}(c_n = k) \log \pi_{nc}
\end{align*}

Dropping terms that don't include $c_n$, the right-hand side (RHS) contains two relevant terms:
\begin{align*}
    &\mathbb{E}_{q(\{\phi_{nc}\})}[\log p(o_n, c_n, \{\phi_{nc}\}|o_{<n})]\\
    &= \mathbb{E}_{q(\{\phi_{nc}\})}[\log p(c_n|o_{<n}) + \log p(o_n|c_n, \{\phi_{nc}\})]\\
    &= \log q(c_n|o_{<n}) + \mathbb{E}_{q(\{\phi_{nc}\})}[\log p(o_n|c_n, \{\phi_{nc}\})]
\end{align*}

The first term is determined by the Dynamical CRP prior:
\[\log q(c_n|o_{<n}) = \sum_k \mathbbm{I}(c_n = k) \log q(c_n = k | o_{<n}) \]

The second term is given by:
\begin{align*}
    &\mathbb{E}_{q(\{\phi_{nc}\})}[\log p(o_n|c_n, \{\phi_{nc}\})]\\
    &= \mathbb{E}_{q(\{\phi_{nc}\})} \Big[ \sum_k  -\frac{1}{2 \sigma_o^2} ||o_n - \phi_{nc}||^2 \, \mathbbm{I}(c_n = k) \Big]\\
    &= \sum_k  -\frac{1}{2 \sigma_o^2} (o_n^T o_n - 2 o_n^T \mu_{nc} + \Tr[\Sigma_{nc} + \mu_{nc} \mu_{nc}^T]) \, \mathbbm{I}(c_n = k)
\end{align*}

Comparing the simplified left-hand and right-hand sides, and solving for the variational parameter of the probability of the $l$th cluster $\pi_{nl}$:
\begin{align*}
    \pi_{nl} &\propto \exp \Big( \log q(c_n = l | o_{<n}) - \frac{1}{2\sigma_o^2} o_n^T o_n + \frac{1}{\sigma_o^2} o_n^T \mu_{nl} - \frac{1}{2\sigma_o^2} \Tr[\Sigma_{nl} + \mu_{nl} \mu_{nl}^T] \Big)
\end{align*}

For the $l$th cluster centroid's variational parameters, we want to solve:
\begin{align*}
    \log q(\phi_{nl} | o_{\leq n}; \mu_{nl}, \Sigma_{nl})
    = \mathbb{E}_{q}[\log p(o_n, c_n, \{\phi_{nc}\}|o_{<n})]
\end{align*}

The LHS is:
\begin{align*}
    &\log q(\phi_{nl} | o_{\leq n}; \mu_{nl}, \Sigma_{nl})\\
    &= - \frac{1}{2} (\phi_{nl} - \mu_{nl})^T \Sigma_{nl}^{-1} (\phi_{nl} - \mu_{nl})
\end{align*}

The RHS is:
\begin{align*}
    \mathbb{E}_{q}[\log p(o_n, c_n, \{\phi_{nc}\}|o_{<n})] 
    &= \mathbb{E}_{q}[\log q(\phi_{nl}|o_{<n}) + \log p(o_n|c_n, \{\phi_{nc}\}, o_{<n})]
\end{align*}

where the first RHS term is:
\begin{align*}
    \mathbb{E}_{q}[\log q(\phi_{nl}|o_{<n})]
    &= -\frac{1}{2} (\phi_{nl} - \mu_{n-1, l})^T \Sigma_{n-1, l}^{-1} (\phi_{nl} - \mu_{n-1, l})
\end{align*}

and the second RHS term is:
\begin{align*}
    \mathbb{E}_{q}[\log p(o_n|c_n, \{\phi_{nc}\}, o_{<n})]
    &=\mathbb{E}_{q} \Big[ \sum_{k} -\frac{1}{2 \sigma_o^2} (o_n - \phi_{nc})^T (o_n - \phi_{nc}) \mathbbm{I}(c_n = k) \Big]\\
    &= -\frac{1}{2 \sigma_o^2} (o_n - \phi_{nl})^T (o_n - \phi_{nl}) \, \pi_{nl}
\end{align*}

Setting the LHS and RHS equal and isolating terms quadratic in $\phi_{nl}$ allows us to solve for the variational covariance parameter:
\begin{align*}
    \phi_{nl}^T \Sigma_{nl}^{-1} \phi_{nl} &=\phi_{nl}^T \Sigma_{n-1, l}^{-1} \phi_{nl} + \phi_{nl}^T \big(\frac{\pi_{nl}}{\sigma_0^2} I \big) \phi_{nl}\\
    \Sigma_{nl} &= \Big(\Sigma_{n-1, l}^{-1} + \frac{\pi_{nl}}{\sigma_0^2} I  \Big)^{-1}
\end{align*}

Isolating terms linear in $\phi_{nl}$ allows us to solve for the variational mean parameter:
\begin{align*}
    \phi_{nl}^T \Sigma_{nl}^{-1} \mu_{nl} &= \phi_{nl}^T \Sigma_{n-1, l}^{-1} \, \mu_{n-1, l} + \phi_{nl}^T (\frac{\pi_{nl}}{\sigma_o^2} I ) o_n\\
    \mu_{nl} &= \Sigma_{nl} \Big( \Sigma_{n-1, l}^{-1} \, \mu_{n-1, l} + \frac{\pi_{nl}}{\sigma_o^2} o_n \Big)\\
\end{align*}


\subsection{CAVI for (Product of) Bernoulli Likelihoods}

    

Mean-field family:

\begin{align*}
    q(c_n, \{\phi \} | o_{\leq n}) \defeq q(c_n|o_{\leq n}; \pi_{n})& \prod_{c=1}^{C_n} q(\phi_{nc}|o_{\leq n}; \gamma_{nc}, \beta_{nc}) \\
    q(c_n|o_{\leq n}; \{\pi_n \}) &\defeq Categorical(\pi_n)\\
    q(\phi_{nc}|o_{\leq n}; \gamma_{nc}, \beta_{nc}) &\defeq \prod_{c} \prod_l Beta(\gamma_{ncl}, \beta_{ncl})
\end{align*}

where $\theta_n \defeq \{\pi_{nc}\}_c \cup \{\gamma_{ncl}\} \cup \{\kappa_{ncl}\}$ are our variational parameters for the $n$th observation. The mixture weights' parameters $\pi_n$ will be determined by solving the following:
\begin{align*}
    \log q(c_n|o_{\leq n}; \pi_{n}) &= \mathbb{E}_{q(\{\phi_{nc}\})}[\log p(o_n, c_n, \{\phi_{nc}\}|o_{<n})]
\end{align*}

The left-hand side (LHS) is:
\begin{align*}
    &\log q(c_n|o_{\leq n}; \pi_{n}) = \sum_c \mathbbm{I}(c_n = c) \log \pi_{nc}
\end{align*}

Dropping terms that don't include $c_n$, the right-hand side (RHS) contains two relevant terms:
\begin{align*}
    \mathbb{E}_{q}[\log p(o_n, c_n, \{\phi_{nc}\}|o_{<n})]
    &= \mathbb{E}_{q}[\log p(c_n|o_{<n}) + \log p(o_n|c_n, \{\phi_{nc}\})]\\
    &= \log q(c_n|o_{<n}) + \mathbb{E}_{q}[\log p(o_n|c_n, \{\phi_{nc}\})]
\end{align*}

The first term is determined by the Dynamical CRP prior:
\[\log q(c_n|o_{<n}) = \sum_c \mathbbm{I}(c_n = c) \log q(c_n = c | o_{<n}) \]

The second term is given by:
\begin{align*}
    &\mathbb{E}_{q(\{\phi_{nc}\})}[\log p(o_n|c_n, \{\phi_{nc}\})]\\
    &= \mathbb{E}_{q(\{\phi_{nc}\})} \Bigg[\log  \prod_c \Big(\prod_l \phi_{ncl}^{x_{nl}} (1 - \phi_{ncl})^{1 - x_{nl}} \Big)^{\mathbbm{I}(c_n = c)}  \Bigg]\\
    &= \mathbb{E}_{q(\{\phi_{nc}\})} \Bigg[\sum_c \mathbbm{I}(c_n = c) \sum_l \log \Big(\phi_{ncl}^{x_{nl}} (1 - \phi_{ncl})^{1 - x_{nl}} \Big)  \Bigg]\\
    &= \sum_c \mathbbm{I}(c_n = c) \sum_l \Big( x_{nl} \mathbb{E}_{q(\phi_{ncl})}[\log \phi_{ncl}]  + (1 - x_{nl}) \mathbb{E}_{q(\phi_{ncl})}[\log (1 - \phi_{ncl})] \Big)\\
    &= \sum_c \mathbbm{I}(c_n = c) \sum_l \Big( x_{nl} (\psi(\gamma_{ncl}) - \psi(\gamma_{ncl} + \beta_{ncl})) + (1 - x_{nl}) (\psi(\beta_{ncl}) - \psi(\gamma_{ncl} + \beta_{ncl})) \Big)
\end{align*}

where $\psi(x) \defeq \frac{d}{dx} \log \Gamma(x)$ is the digamma function.  Comparing the simplified left-hand and right-hand sides, and solving for the variational parameter of the probability of the $l$th cluster $\pi_{nl}$:
\begin{align*}
    \pi_{nc} &\propto \exp \Bigg( \log q(c_n = c | o_{<n}) + \sum_l \Big( x_{nl} (\psi(\gamma_{ncl}) - \psi(\gamma_{ncl} + \beta_{ncl}))\\
    &\quad \quad \quad \quad \quad + (1 - x_{nl}) (\psi(\beta_{ncl}) - \psi(\gamma_{ncl} + \beta_{ncl})) \Big) \Bigg)
\end{align*}

For the $c$-th cluster's variational parameters, we want to solve:
\begin{align*}
    \log q(\phi_{ncl} | o_{\leq n}; \gamma_{ncl}, \beta_{ncl})
    = \mathbb{E}_{q}[\log p(o_n, c_n, \{\phi_{nc}\}|o_{<n})]
\end{align*}

The LHS is:
\begin{align*}
    \log q(\phi_{ncl} | o_{\leq n}; \gamma_{ncl}, \beta_{ncl})& = \log Beta(\phi_{ncl}; \gamma_{ncl}, \beta_{ncl})\\
    &= (\gamma_{ncl} - 1 ) \log (\phi_{ncl}) + (\beta_{ncl} - 1) \log (1 - \phi_{ncl})
\end{align*}

Dropping terms that don't contain $\phi_{ncl}$, the RHS is:
\begin{align*}
    &\mathbb{E}_{q(-\phi_{ncl})}[\log p(o_n, c_n, \{\phi_{nc}\}|o_{<n})]\\
    &=\mathbb{E}_{q(-\phi_{ncl})} \Bigg[\sum_{c'} \mathbbm{I}(c_n = c') \sum_{l'} \log \Big(\phi_{ncl'}^{x_{nl'}} (1 - \phi_{ncl'})^{1 - x_{nl'}} \Big)  \Bigg]+ \mathbb{E}_{q(-\phi_{ncl})} \Bigg[ q(\phi_{ncl}|o_{<n})\Bigg]\\
    &= \pi_{nc} \Big( x_{nl} \log (\phi_{ncl}) + (1 - x_{nl}) \log (1 - \phi_{ncl}) \Big) + (\gamma_{n-1,cl} - 1) \log (\phi_{ncl}) + (\beta_{n-1, cl} - 1 ) \log (1 - \phi_{ncl})
\end{align*}

Grouping terms and setting equal:
\begin{align*}
    \gamma_{ncl} - 1 &= \pi_{nc} x_{nl} + \gamma_{n-1, cl} - 1\\
    \gamma_{ncl} &= \pi_{nc} x_{nl} + \gamma_{n-1, cl}\\
    \beta_{ncl} - 1 &= \pi_{nc} (1 - x_{nl}) + \beta_{n-1, cl} - 1\\
    \beta_{ncl} &= \pi_{nc} (1 - x_{nl}) + \beta_{n-1, cl}\\
\end{align*}

\subsection{CAVI for von-Mises-Fisher Likelihood}

Mean field family:
\begin{align*}
    q(c_n, \{\phi \} | o_{\leq n}) \defeq q(c_n|o_{\leq n}; \{\pi_{nc}\})& \prod_{k=1}^{C_n} q(\phi_{nc}|o_{\leq n}; \mu_{nc}, \Sigma_{nc}) \\
    q(c_n|o_{\leq n}; \{\pi_{nc} \}) &\defeq Categorical(\pi_n)\\
    q(\phi_{nc}|o_{\leq n}; \mu_{nc}, \kappa_{nc}) &\defeq \mathcal{VMF}(\mu_{nc}, \kappa_{nc})
\end{align*}

where $\theta_n \defeq \{\pi_{nc}\}_k \cup \{\mu_{nc}\}_k \cup \{\kappa_{nc}\}_k$ are our variational parameters for the $n$th observation. The mixture weights' parameters $\pi_n$ will be determined by solving the following:

\begin{align*}
    \log q(c_n|o_{\leq n}; \pi_{n}) &= \mathbb{E}_{q(\{\phi_{nc}\})}[\log p(o_n, c_n, \{\phi_{nc}\}|o_{<n})]
\end{align*}

The left-hand side (LHS) is:
\begin{align*}
    &\log q(c_n|o_{\leq n}; \pi_{n}) = \sum_k \mathbbm{I}(c_n = k) \log \pi_{nc}
\end{align*}

Dropping terms that don't include $c_n$, the right-hand side (RHS) contains two relevant terms:
\begin{align*}
    \mathbb{E}_{q}[\log p(o_n, c_n, \{\phi_{nc}\}|o_{<n})]
    &= \mathbb{E}_{q}[\log p(c_n|o_{<n}) + \log p(o_n|c_n, \{\phi_{nc}\})]\\
    &= \log q(c_n|o_{<n}) + \mathbb{E}_{q}[\log p(o_n|c_n, \{\phi_{nc}\})]
\end{align*}

The first term is determined by the RNCRP prior:
\[\log q(c_n|o_{<n}) = \sum_k \mathbbm{I}(c_n = k) \log q(c_n = k | o_{<n}) \]

The second term is given by:
\begin{align*}
    \mathbb{E}_{q(\{\phi_{nc}\})}[\log p(o_n|c_n, \{\phi_{nc}\})]
    &= \mathbb{E}_{q(\{\phi_{nc}\})} \Big[ \sum_k  \frac{1}{\sigma_o^2} \, \phi_{nc}^T \, o_n \, \mathbbm{I}(c_n = k) \Big]\\
    &= \sum_k  \frac{1}{\sigma_o^2} \, \left(\frac{I_{D/2 - 1}^{\prime} (\kappa)}{I_{D/2 - 1} (\kappa)} - \frac{D/2 - 1}{\kappa} \right) \mu_{nc}^T o_n   \, \mathbbm{I}(c_n = k)
\end{align*}

Comparing the simplified left-hand and right-hand sides, and solving for the variational parameter of the probability of the $l$th cluster $\pi_{nl}$:
\begin{align*}
    \pi_{nl} &\propto \exp \Bigg( \log q(c_n = l | o_{<n}) + \frac{1}{\sigma_o^2} \Bigg( \frac{I_{D/2 - 1}^{\prime} (\kappa)}{I_{D/2 - 1} (\kappa)} - \frac{D/2 - 1}{\kappa} \Bigg) \mu_{nc}^T o_n \Bigg)
\end{align*}

For the $l$th cluster centroid's variational parameters, we want to solve:
\begin{align*}
    \log q(\phi_{nl} | o_{\leq n}; \mu_{nl}, \kappa_{nl})
    = \mathbb{E}_{q}[\log p(o_n, c_n, \{\phi_{nc}\}|o_{<n})]
\end{align*}

The LHS is:
\begin{align*}
    \log q(\phi_{nl} | o_{\leq n}; \mu_{nl}, \kappa_{nl}) &= \kappa_{nl} \, \mu_{nl}^T \, \phi_{nl}
\end{align*}

The RHS is:
\begin{align*}
    \mathbb{E}_{q}[\log p(o_n, c_n, \{\phi_{nc}\}|o_{<n})]
    &= \mathbb{E}_{q}[\log q(\phi_{nl}|o_{<n}) + \log p(o_n|c_n, \{\phi_{nc}\}|o_{<n})]
\end{align*}

where the first RHS term is:
\begin{align*}
    \mathbb{E}_{q}[\log q(\phi_{nl}|o_{<n})] = \kappa_{n-1,l}\, \mu_{n-1, l}^T \, \phi_{nl}
\end{align*}

and the second RHS term is:
\begin{align*}
    \mathbb{E}_{q}[\log p(o_n|c_n, \{\phi_{nc}\}, o_{<n})]
    &=\mathbb{E}_{q} \Big[ \sum_{k} \frac{1}{\sigma_o^2} \, \phi_{nc}^T \, o_n \, \mathbbm{I}(c_n = k) \Big] = \frac{1}{\sigma_o^2} \phi_{nl}^T \, o_n \, \pi_{nl}
\end{align*}

Setting the LHS and RHS equal:

\[\kappa_{nl} \, \mu_{nl} =  \kappa_{n-1, l} \, \mu_{n-1, l} + \frac{1}{\sigma_o^2} \pi_{nl} \, o_n\]

The two variational parameters are separately recoverable by setting $\mu_{nl}$ equal to the unit direction of the right hand side $RHS \defeq \kappa_{n-1, l} \, \mu_{n-1, l} + \frac{1}{\sigma_o^2} \pi_{nl} \, o_n$ and by setting $\kappa_{nl}$ equal to the magnitude of the right hand side:

\[\kappa_{nl} = \lvert \lvert RHS \rvert \rvert_2 \quad \quad \mu_{nl} = \frac{RHS}{\lvert \lvert RHS \rvert \rvert_2}\]




